\documentclass[preprint,12pt]{elsarticle}

\usepackage{mathrsfs}
\usepackage{blindtext}
\usepackage{algorithm}
\usepackage{algpseudocode}
\usepackage{graphicx}
\usepackage[utf8]{inputenc}
\usepackage{lmodern}
\usepackage[T1]{fontenc}
\usepackage{lscape}
\usepackage{amsmath}	
\usepackage{amssymb}
\usepackage{cases}
\usepackage{graphicx}
\usepackage{caption}
\usepackage{comment}

\usepackage{makecell}
\usepackage{multirow}
\usepackage{amsfonts}
\usepackage{float} 
\usepackage{diagbox}
\usepackage{fullpage}
\usepackage{url}
\usepackage{rotating}
\usepackage{eurosym}
\usepackage{wrapfig}
\usepackage[final]{pdfpages} 
\usepackage{epstopdf} 
\usepackage[a4paper]{geometry}
\usepackage[nottoc]{tocbibind}
\usepackage{placeins}
\usepackage{float}
\usepackage{listings}
\usepackage{color, colortbl}
\usepackage{hyperref}
\usepackage{xcolor}
\usepackage{graphicx, caption, subcaption}

\usepackage{sidecap}
\hypersetup{
colorlinks,
citecolor=black,
filecolor=black,
linkcolor=black,
urlcolor=black
}

\journal{Computer Methods in Applied Mechanics and Engineering}
\graphicspath{{./figures/}}

\begin{document}

\begin{frontmatter}


\title{Multi-fidelity physics constrained neural networks for dynamical systems}

\author[inst1]{Hao Zhou}

\affiliation[inst1]{organization={Department of Earth Science \& Engineering, Imperial College London },
            country={UK}}

\author[inst2]{Sibo Cheng\corref{cor1}}

\author[inst1]{Rossella Arcucci}


\cortext[cor1]{corresponding: sibo.cheng@imperial.ac.uk}

\affiliation[inst2]{organization={Data Science Institute, Department of Computing, Imperial College London },
country={UK}}

\begin{figure}
    \begin{center}
    \includegraphics[width=1.\textwidth]{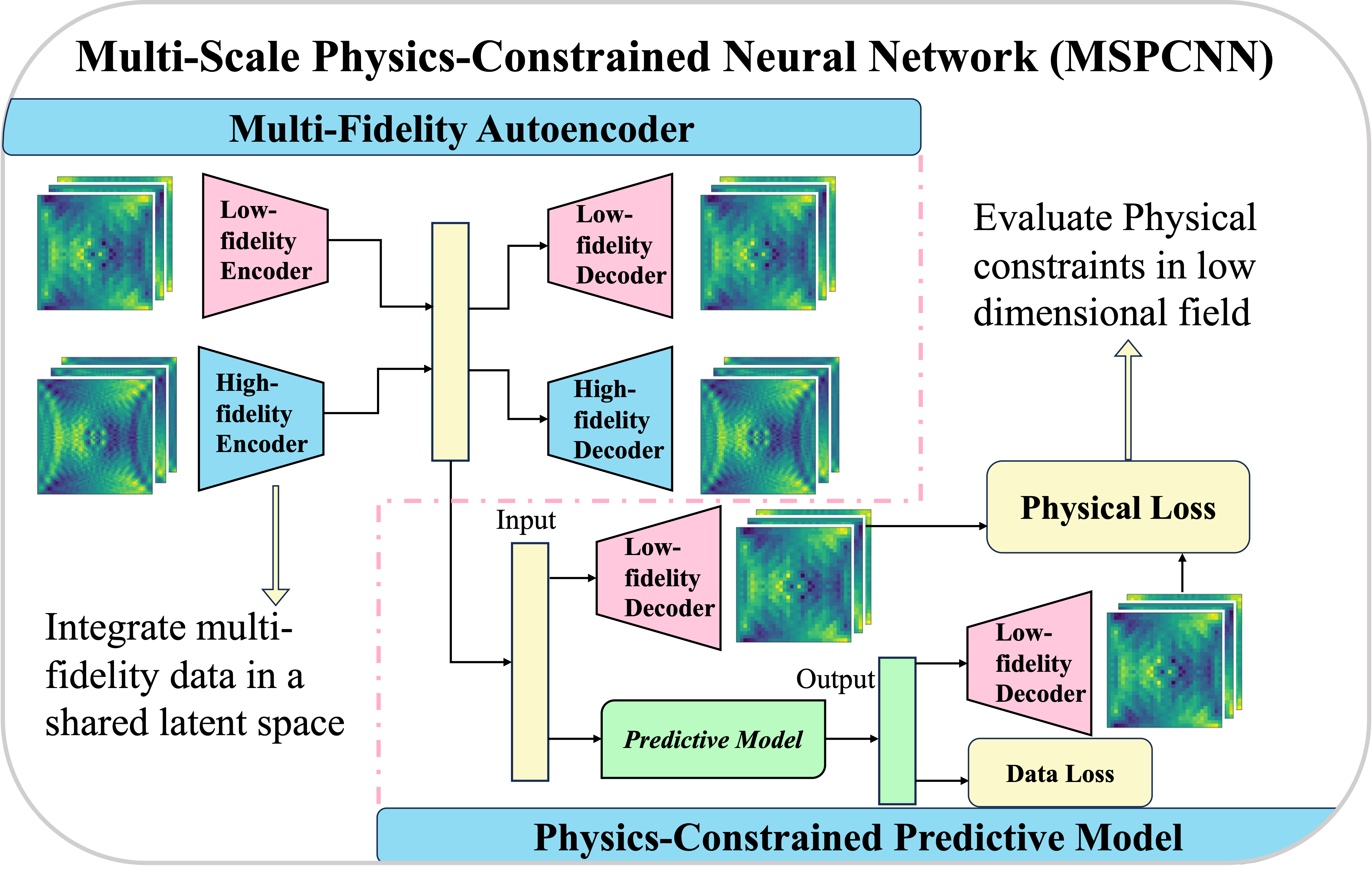}
    \end{center}
\captionsetup{labelformat=empty}
\caption{Graphical Abstract}
\end{figure}

\begin{abstract}
Physics-constrained neural networks are commonly employed to enhance prediction robustness compared to purely data-driven models, achieved through the inclusion of physical constraint losses during the model training process. However, one of the major challenges of physics-constrained neural networks consists of the training complexity especially for high-dimensional systems. In fact, conventional physics-constrained models rely on singular-fidelity data necessitating the assessment of physical constraints within high-dimensional fields, which introduces computational difficulties. Furthermore, due to the fixed input size of the neural networks, employing multi-fidelity training data can also be cumbersome. In this paper, we propose the Multi-Scale Physics-Constrained Neural Network (MSPCNN), which offers a novel methodology for incorporating data with different levels of fidelity into a unified latent space through a customized multi-fidelity autoencoder. Additionally, multiple decoders are concurrently trained to map latent representations of inputs into various fidelity physical spaces. As a result, during the training of predictive models, physical constraints can be evaluated within low-fidelity spaces, yielding a trade-off between training efficiency and accuracy. In addition, unlike conventional methods, MSPCNN also manages to employ multi-fidelity data to train the predictive model. We assess the performance of MSPCNN in two fluid dynamics problems, namely a two-dimensional Burgers’ system and a shallow water system. Numerical results clearly demonstrate the enhancement of prediction accuracy and noise robustness when introducing physical constraints in low-fidelity fields. On the other hand, as expected, the training complexity can be significantly reduced by computing physical constraint loss in the low-fidelity field rather than the high-fidelity one.
\end{abstract}

\begin{keyword}
Reduced-order modelling \sep Multiple fidelity \sep Physical constraints \sep LSTM networks \sep Dynamical systems \sep Long-time prediction


\end{keyword}

\end{frontmatter}
\newgeometry{top=1.5cm}
\section*{Main Notations} 
\begin{tabular}{ll}
\hline
\textbf{Notation} & \textbf{Description} \\
\hline
\\
  & \textit{Multi-Scale Physics-Constrained Neural Network} \\
  \\
$\mathit{\mathbf{x}}_{t}$ & State vector in the full space at time t \\
$\boldsymbol{\eta}_{t}$ & Compressed state vector in the latent space at time t\\
$\mathit{\mathbf{x}}^r_{t}$ & Reconstruction state vector in the full space at time t \\
$\mathcal{F}_e, \mathcal{F}_d$ & Encoder, Decoder function in autoencoder\\
$\theta_{\mathcal{F}_e}, \theta_{\mathcal{F}_d}$ & Parameters for the encoder and decoder \\
$\textrm{N}_\textrm{step}$ & Total number of time steps in dataset \\
$k_\textrm{in}, k_\textrm{out}$ & Input and Output time steps of LSTM \\
$\tilde{\boldsymbol{\eta}}_t$ & Output of LSTM in the latent space at time t\\
${\boldsymbol{\eta}}_{t:t+k_\textrm{in}-1}$ & Sequence of compressed state vectors \\
$\mathcal{J}$ & Loss function \\
$l_\textrm{data}$ & Loss between the predicted and true latent representations\\
$l_\textrm{physics}$ & General loss of physical constraint \\
$\alpha$ & Coefficient of physical loss \\
$l_\textrm{energy}, l_\textrm{flow}$ & Loss of energy conservation, loss of flow operator\\
$\mathcal{F}_{LSTM}, \theta_{\textrm{LSTM}}$ & LSTM function, Parameters of LSTM\\
$E_\textrm{in}, E_\textrm{out}$ & Total energy of input and output sequence \\
$\mathcal{E}$ & Function to calculate total energy \\
$f$ & Flow operator function \\
$\mathbf{x}^\textrm{fp}$ & State vector predicted by flow operator \\
$\mathbf{X}_{h,\textrm{train}}, \mathbf{X}_{l,\textrm{train}}$ & High and low-fidelity datasets \\
$\mathbf{x}_{h,t}, \mathbf{x}_{h,t}^r$ & Original and Reconstructed high-fidelity data \\
$\mathcal{F}_{h,e}, \mathcal{F}_{h,d}$ & Encoder and Decoder of high-fidelity data \\
$\mathbf{x}_{l,t}, \mathbf{x}_{l,t}^r$ & Original and Reconstructed low-fidelity data \\
$\mathcal{F}_{l,e}, \mathcal{F}_{l,d}$ & Encoder and Decoder of low-fidelity data \\
$\mathit{\eta}_{l,j}$ & Compressed low-fidelity data in latent space \\
$\mathbf{x}_{l}^\textrm{fp}$ & State vector predicted by flow operator in low-fidelity field \\
\\
  & \textit{2D Burgers' equation test case} \\
  \\
$u, v$ & Velocity components in the x (horizontal) and y (vertical) directions \\
$t$ & time \\
$x, y$ & Corrdinate system \\
$Re$ & Reynolds number \\
\\
  & \textit{Shallow water equation test case} \\
  \\
$h$ & Total water depth including the undisturbed water depth\\
$u, v$ & Velocity components in the x (horizontal) and y (vertical) directions\\
$g$ & Gravitational acceleration\\
$r$ & Spatial euclidean distance\\
$\epsilon$ & Balgovind type of correlation function\\
$L$ & Typical correlation length scale\\
\hline
\end{tabular}
\restoregeometry
\section{Introduction}
Computational simulations of fluids and other complex physical systems have critical applications in engineering and the physical sciences such as aerodynamics~\cite{tabatabaei_techniques_2022}, heat transfer~\cite{en12214156} and acoustics~\cite{acoustics4040056}. Historically, many of these systems were effectively described using partial differential equations (PDEs). Traditional discretization and solution approaches, such as Finite Difference Method~\cite{casulli_semi-implicit_1990,kurganov_central-upwind_2002}, Finite Volume Method~\cite{alcrudo_high-resolution_1993, bale2003wave} and Lattice Boltzmann Method~\cite{qian_lattice_1992, shan_lattice_1993}, have been proven reliable for achieving high fidelity and high accuracy results. However, the slow computational speed and demanding significant resources~\cite{ babanezhad_functional_2020, lagha_body_2023} make it less ideal for real-time predictions in high dimensional systems. When conducting simulations of transient smoke or pollutant transport within an enclosed space, such as a hotel lobby, conventional computational fluid dynamics (CFD) techniques can require a full day of computational time on a personal computer for a just 10-minute event~\cite{zuo2009real}.

Faced with the high computational demands of traditional fluid dynamics methods~\cite{berkooz1993proper,mohan2018deep, kingma2013auto}, researchers increasingly turn to Reduced Order Modeling (ROM), encompassing deep learning (DL) and machine learning (ML) technologies~\cite{fresca2021real, drakoulas2023fastsvd}. Autoencoders (AE) and recurrent neural networks (RNN) such as Long Short-Term Memory (LSTM)~\cite{hochreiter1997long} networks are especially important in this regard, used for efficiently processing data and predicting evolution in latent space. For instance, Maulik et al.~\cite{maulik_reduced-order_2021} employed a convolutional autoencoder (CAE) combined with LSTM to address the shortcomings of the proper orthogonal decomposition (POD) in capturing interactions during temporal evolution. Building on this, Nakamura et al.~\cite{nakamuracnn} introduced a CAE-LSTM model for high-dimensional turbulent channel flow systems. Meanwhile, Kim et al.~\cite{kim2019deep} adopted a convolutional neural network (CNN) based generative model for parameterized fluid velocity fields, streamlining both fluid simulation and data compression. However, these purely data-driven methods face challenges, particularly in ensuring generalisation capability for new scenarios~\cite{kissas2020machine} and guaranteeing physically realistic outputs~\cite{wang2004image,mohan_spatio-temporal_2020,wu2023deep}. 

To address these issues, Physics-Constrained Neural Networks (PCNN)~\cite{raissi_physics-informed_2019, karniadakis_physics-informed_2021,qu_can_2022} improve model accuracy and generalisation ability by introducing physical constraint losses during the training process. PCNN integrates physical constraints into the model, reducing dependency on large amounts of high-quality training data, guiding optimisation paths, improving generalisation errors, and reducing prediction uncertainty~\cite{nghiem_physics-informed_2023, yang2023physics}. For instance, Fu et al.~\cite{fu2023physics} introduce a Physics-Data Combined Machine Learning (PDCML) approach that employs Proper Orthogonal Decomposition (POD) and physical constraints to enhance parametric reduced-order modeling, particularly in limited data contexts. Mohan et al.~\cite{mohan_embedding_2023} proposed a CNN model that incorporates the incompressibility of a fluid flow and demonstrated its effectiveness. Karbasian et al.~\cite{karbasian_application_2022} developed a new approach for PDE-constrained optimisation of nonlinear systems that transformed the physical equations from physical space to non-physical space. In the prototype problem of fluid flow prediction, Erichson et al.~\cite{erichson_physics-informed_2019} proposed a model that incorporates physical information constraints and maintains Lyapunov stability by training an AE, which not only improves generalisation error but also reduces prediction uncertainty.

Although incorporating physical constraints into machine learning offers numerous advantages over purely data-driven approaches, it comes with its own set of challenges. During the training of ROMs, the direct application of the physical laws isn't straightforward as the evolution transpires in latent space. The latent representations need to be decoded from the latent space back to the full physical space to evaluate these laws~\cite{chen2021physics}. However, due to the fixed input size of the ROMs, especially when inputs are in high-fidelity field, employing physical constraints will consume a lot of computing resources. Therefore, if we can map the latent space driven from a high-fidelity field to a low-fidelity counterpart, the physical constraints can be applied within the low-fidelity space. By doing so, we unlock the potential to leverage the physical constraint losses at a low-fidelity level for model optimisation, effectively alleviating the computational burdens and complexities. Moreover, in real-world scenarios, we often encounter data in varying fidelities, which cannot be fully used due to the fixed neural network input size. Examples could be found in the field of meteorology\cite{zhang2022multi,gao2022multi,li2022graph}. The data is obtained from several sources, including ground stations, satellites, balloons, and aircraft, each offering information with varying degrees of accuracy and reliability. Ground stations provide data that is specific to a particular location, whereas satellites offer a wider coverage area but with a decreased level of detail~\cite{de2023spatial}. As a result of limitations in model input size, it is hard to fully leverage all of the multi-fidelity data. Besides, low-fidelity data is easier and cheaper to obtain, while high-fidelity data is more resource consuming~\cite{conti2023multi}. If our high-fidelity data and its low-fidelity counterpart can achieve the same latent representation, an anticipated method would efficiently leverage all the levels of data fidelity for training and guide and constrain the high-fidelity modelling by low-fidelity physical constraints, ensuring a balance between computational efficiency and physical accuracy.

In recent years, multi-fidelity data has been harnessed primarily for several central purposes. Firstly, a surrogate model will be employed to integrate models trained on data of varying fidelity, aiming to construct a comprehensive model that captures the accuracy of high-fidelity data and the computational efficiency of low-fidelity data. Xiong et al.~\cite{xiong2007new} proposed a model fusion technique based on Bayesian-Gaussian process modeling to develop cost-effective surrogate models, integrating data from both high-fidelity and low-fidelity sources and quantifying the surrogate model's interpolation uncertainty. Secondly, it involves utilising low-fidelity data to estimate or generate high-fidelity data, hence circumventing the computational expenses associated with directly obtaining high-fidelity data through simulations. Geneva et al.~\cite{geneva2020multi} provide a multi-fidelity deep generative model that is specifically developed for surrogate modelling of turbulent flow fields with high-fidelity utilising data obtained from a low-fidelity solver. In addition, multi-fidelity data is used to fine-tune the varying parameters in multi-scale PDEs to enhance predictive accuracy. Park et al.~\cite{park2022physics} proposed an approach that adopted a physics-informed neural network that leverages a priori knowledge of the underlying homogenised equations to estimate model parameters based on multi-scale solution data. Finally, there is an emerging practice of utilising low-fidelity data as an additional resource to improve the effectiveness of high-fidelity models. Romor et al.~\cite{romor2021multi} constructed a low-fidelity response surface based on gradient-based reduction, which facilitates the updating of the nonlinear autoregressive multi-fidelity Gaussian process. However, to the best of the author's knowledge, there is no existing model or method that can leverage physical constraints in low-fidelity field to both alleviate computational burdens and ensure prediction accuracy. 

In response to the above challenges, we introduce a deep learning method designed for multi-scale physical constraints, termed the Multi-Scale Physics-Constrained Neural Network (MSPCNN). Our methodology involves employing two distinct AE models tailored for high- and low-fidelity data, respectively. The first AE is trained exclusively on the high-fidelity data. For the second AE, we separately train its encoder on low-fidelity data to map it into the same latent space as the first AE, and its decoder to reconstruct the low-fidelity data from the latent representations driven from high-fidelity counterparts. Subsequently, we formulate an LSTM model embedded with physical constraints that takes the latent representations obtained by the AEs as input and uncovers the evolution laws of the physical system within the latent space. During the training of the LSTM, besides the basic metrics, such as MSE, the compressed data will be decoded to the low-fidelity field, forming the computation of the physical constraint loss that guides model refinement. Additionally, because the LSTM accepts the latent representations as input, which can be derived from data in various fidelities, the low-fidelity data can contribute to the training of high-fidelity surrogate models, considerably curbing its computational demands~\cite{yu2019non}. In our study, we selected two numerical tests, a two-dimensional Burgers' system and a Shallow Water system. Both of these cases are frequently employed as benchmarks in scientific machine learning~\cite{cheng2019background, maulik2021latent, liu2022enkf}. Specifically, the Burgers' system is characterized by its relative simplicity and its ability to depict two-dimensional variations in viscous fluids. Conversely, the Shallow Water system captures the two-dimensional horizontal dynamics of a body of water. Moreover, the Shallow Water equations encompass several temporal and spatial scales, rendering it well-suited for the validation of multi-scale models like MSPCNN.

In summary, we make the following contributions in this study:
\begin{enumerate}
    \item We propose a novel physics-contrained machine learning model, named MSPCNN. It innovatively leverages physical constraints in low-fidelity field for the training of high-fidelity models, making a balance between computational efficiency and physical accuracy.
    \item By integrating and unifying data of varying fidelity, multi-fidelity data can be used for training MSPCNN. This integration also ensures that the trained models can be flexibly adapted to yield results across different fidelity levels.
    \item MSPCNN demonstrates robust performance in the presence of noisy data as compared with conventional PCNN.
    \item MSPCNN is rigorously tested on two CFD models. Compared to the ROMs without physical constraints, the proposed MSPCNN with multiple physical constraints demonstrates a significant reduction in MSE by at least 50\%. Furthermore, in terms of training time, compared against high-fidelity physics-constrained neural netowrks, MSPCNN exhibits a remarkable reduction, ranging from half to a quarter of the original computation time.
\end{enumerate}

The rest of this paper is organised as follows. In Section~\ref{sec:PCNN}, we introduce the state-of-the-art PCNNs for high-dimensional dynamical systems. Section~\ref{sec:MSPCNN} presents the structure of MSPCNN and details the training methodology for it. Two numerical experiments, specifically a two-dimensional Burgers’ system and a Shallow Water system, are discussed in Section~\ref{sec:burgers} and Section~\ref{sec:sw}, respectively. Finally, we conclude and summarise our findings in Section~\ref{sec:conclusion}.

\section{Physics constrained reduced order modelling: state of the art}
\label{sec:PCNN}
\begin{figure}
    \begin{center}
        \includegraphics[width=0.8\textwidth]{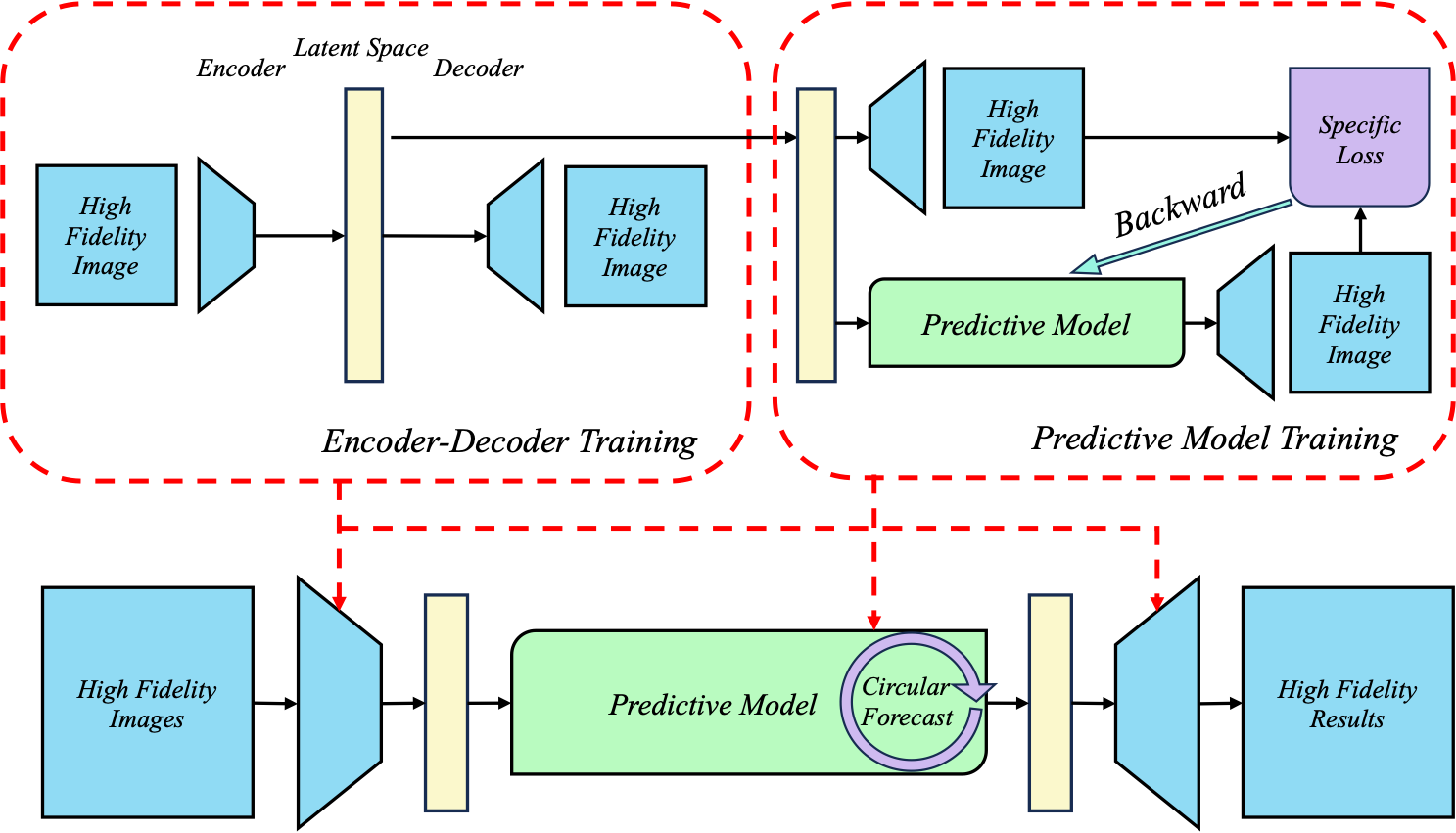}
    \end{center}
    \caption{\label{fig:LSTM_flowchart} Flowchart of PCNN}
\end{figure}

This section focuses on the structure of state-of-art PCNNs for high dimensional dynamic systems. These models include reduced order modelling (AE), surrogate models based on recurrent neural networks (LSTM), and the incorporation of physical constraints and they are integrated in the way as shown in Fig.~\ref{fig:LSTM_flowchart}~\cite{mohan_embedding_2023, conti2023multi}.
\subsection{Reduce Order Modelling: AE}
An AE is a specialised form of neural network designed to reduce the dimensionality of input data while preserving its key features.

AE operates through an encoder-decoder architecture as shown in Fig.~\ref{fig:LSTM_flowchart} Encoder-Decoder Training part. The encoder $\mathcal{F}_e$ compresses the input data $\mathit{\mathbf{x}}_{t} = [x_1,x_2,\ldots,x_n] \in \mathbb{R}
^n$ at time $t$ by applying hidden layers and down-sampling, capturing essential features in a compressed latent representation $\boldsymbol{\eta}_{t}= [\eta_1,\eta_2,\ldots,\eta_m]\in \mathbb{R}
^m, m<n$. In contrast, the decoder $\mathcal{F}_d$ works to reconstruct the state vector  $\mathit{\mathbf{x}}^r_{t} = [x_1^r,x_2^r,\ldots,x_n^r] \in \mathbb{R}
^n$ from this latent form $\boldsymbol{\eta}_{t}$, employing up-sampling and hidden layers, i.e.,

\begin{equation}
    \label{eq:CAE}
    \boldsymbol{\eta}_{t} = \mathcal{F}_e(\mathbf{x}_{t}) \quad \text{and} \quad \mathit{\mathbf{x}}^r_{t} = \mathcal{F}_d(\boldsymbol{\eta}_{t})
\end{equation}
The encoder and decoder are trained jointly. The training objective is to minimize the reconstruction error, i.e., the mismatch between the original input and the decoded output. For instance, if we employ the MSE as our loss function $\mathcal{J}(.)$:

\begin{equation}
    \label{eq:CAE_loss}
    \mathcal{J}(\theta_{\mathcal{F}_e}, \theta_{\mathcal{F}_d}) = \frac{1}{\textrm{N}_\textrm{step}}\sum_{i=1}^{\textrm{N}_\textrm{step}} \parallel \mathit{\mathbf{x}}^r_{i} - \mathbf{x}_{i} \parallel_2^2 
\end{equation}
where $\theta_{\mathcal{F}_e} $ and $\theta_{\mathcal{F}_d}$ are the parameters of encoder and decoder, $\{{\mathbf{x}}_{1},{\mathbf{x}}_{2},\ldots,{\mathbf{x}}_{\textrm{N}_\textrm{step}}\}$ representing the total evolution process from initial state to the final state. $\textrm{N}_\textrm{step}$ is the total number of time steps (i.e., training samples), and $\parallel \cdot \parallel_2$ represents Euclidean norm. 
\subsection{RNN-based Surrogate Model: LSTM}
After processing the original data $\mathit{\mathbf{x}}_{t}$ through the AE, the compressed data $\boldsymbol{\eta}_{t}$ in the latent space is obtained. As the next step, it's crucial to understand the dynamics and evolution patterns within these latent representations to make accurate predictions. Since our aim is to predict the physical system behavior in long term, it is essential to choose a model that can efficiently capture temporal dependencies spanning across lengthy sequences. In light of this, researchers have opted for LSTM networks~\cite{xayasouk2020air}. Unlike traditional RNNs, which often struggle with the vanishing gradient problem~\cite{hochreiter1997long}, LSTMs are specifically designed to remember long-range dependencies in sequential data, making them an optimal choice for our requirements. LSTM also delivers a way for sequence to sequence (seq2seq) prediction (LSTM accepts $k_\textrm{in}$ time steps as input and gives $k_\textrm{out}$ time steps as output), which can decrease the online computation time, and more importantly, reduce the accumulated prediction error. For time series that encode latent representations $[{\boldsymbol{\eta}}_{1},{\boldsymbol{\eta}}_{2},\ldots,{\boldsymbol{\eta}}_{\textrm{N}_\textrm{step}}]$, LSTMs can be trained by shifting the starting time step:

\begin{align}
\label{eq:LSTM_Train}
    [{\boldsymbol{\eta}}_{1},\ldots,{\boldsymbol{\eta}}_{k_{in}}] &\xrightarrow{\textrm{Predictive \quad Model 
  \quad Training}} [\tilde{\boldsymbol{\eta}}_{k_\textrm{in}+1}, \ldots, \tilde{\boldsymbol{\eta}}_{k_\textrm{in}+k_\textrm{out}}] \notag \\
      [{\boldsymbol{\eta}}_{2},\ldots,{\boldsymbol{\eta}}_{k_\textrm{in}+1}] &\xrightarrow{\textrm{Predictive \quad Model 
      \quad Training}} [\tilde{\boldsymbol{\eta}}_{k_{in}+2}, \ldots, \tilde{\boldsymbol{\eta}}_{k_\textrm{in}+k_\textrm{out}+1}] \notag \\
      & \vdots \notag \\
      [{\boldsymbol{\eta}}_{\textrm{N}_\textrm{step}-k_\textrm{in}-k_\textrm{out}+1},\ldots,{\boldsymbol{\eta}}_{\textrm{N}_\textrm{step}-k_\textrm{out}}] &\xrightarrow{\textrm{Predictive \quad Model \quad Training}} [\tilde{\boldsymbol{\eta}}_{\textrm{N}_\textrm{step}-k_\textrm{out}+1}, \ldots, \tilde{\boldsymbol{\eta}}_{\textrm{N}_\textrm{step}}]
\end{align}
where $\tilde{\boldsymbol{\eta}}_t$ is the predictive result. During the training phase, various loss functions, such as MSE or mean absolute error (MAE), can be employed to quantify the difference between the predicted latent representations and the true latent representations. When making predictions, we employ it in a circular forecasting to achieve long-time predicting as presented in Fig.~\ref{fig:LSTM_flowchart} and Eq.~\ref{eq:LSTM_Predict}:

\begin{align}
\label{eq:LSTM_Predict}
    [{\boldsymbol{\eta}}_{1},{\boldsymbol{\eta}}_{2},\ldots,{\boldsymbol{\eta}}_{k_\textrm{in}}] &\xrightarrow{\textrm{Predictive \quad Model 
  \quad Prediction}} [\tilde{\boldsymbol{\eta}}_{k_\textrm{in}+1}, \tilde{\boldsymbol{\eta}}_{k_\textrm{in}+2}, \ldots, \tilde{\boldsymbol{\eta}}_{k_\textrm{in}+k_\textrm{out}}] \notag \\
    [\tilde{\boldsymbol{\eta}}_{k_\textrm{in}+1},\ldots,\tilde{\boldsymbol{\eta}}_{k_\textrm{in}+k_\textrm{out}}] &\xrightarrow{\textrm{Predictive \quad Model 
      \quad Prediction}} [\tilde{\boldsymbol{\eta}}_{k_\textrm{in}+k_\textrm{out}+1},  \ldots, \tilde{\boldsymbol{\eta}}_{k_\textrm{in}+2k_\textrm{out}}]  \\
      & \vdots \notag
\end{align}

\subsection{Physical Constraints}
As pointed out by \cite{cheng2023generalised}, reducing the accumulated prediction error becomes especially critical when we use recurrent forecasting to achieve long-time predictions.

The adoption of physical constraints helps to enhance the accuracy and reliability of predictions, which is an important tool for optimising long time forecasts~\cite{cai_physics-informed_2022}. Specifically, ML or DL models can integrate physical constraints by establishing learning biases, which are enforced during the learning process by imposing suitable penalties. Traditionally, the physical constraints can only be applied in the full physical space. Therefore, the latent representations need to be decoded to physical space for evaluating physical loss during the training procedure as shown in Fig.~\ref{fig:LSTM_flowchart} predictive model training part. In a seq2seq prediction model, the composite physics-constrained loss function for a single prediction step, $\mathcal{J}$ (referred to as Specific Loss in Fig.~\ref{fig:LSTM_flowchart}), is given by:

\begin{gather}
    [{\boldsymbol{\eta}}_{t:t+k_\textrm{in}-1}] \xrightarrow{\textrm{Predictive \quad Model 
  \quad Training}} [\tilde{\boldsymbol{\eta}}_{t+k_\textrm{in}:t+k_\textrm{in}+k_\textrm{out}-1}] \notag \\
    \begin{aligned}
        \hspace{-1cm}\mathcal{J}(\theta_{\textrm{LSTM}}) &=  l_\textrm{data}([{\boldsymbol{\eta}}_{t+k_\textrm{in}:t+k_\textrm{in}+k_\textrm{out}-1}], [\tilde{\boldsymbol{\eta}}_{t+k_\textrm{in}:t+k_\textrm{in}+k_\textrm{out}-1}]) \\
        &\hspace{1cm} + \sum_j^c \alpha_j l_\textrm{physics}^j([{\boldsymbol{\eta}}_{t:t+k_\textrm{in}-1}], [\tilde{\boldsymbol{\eta}}_{t+k_\textrm{in}:t+k_\textrm{in}+k_\textrm{out}-1}])
    \end{aligned}
    \label{eq:Loss_Funtion_with_PC}
\end{gather}

where $[{\boldsymbol{\eta}}_{t:t+k_\textrm{in}-1}]$ is the sequence input of LSTM and $[\tilde{\boldsymbol{\eta}}_{t+k_\textrm{in}:t+k_\textrm{in}+k_\textrm{out}-1}] $ is the sequence output of LSTM, $l_\textrm{data} $ denotes the loss function used to measure the discrepancy between the predicted and true latent representations, $l_\textrm{physics}$ represents physics-based regularisation term, $c$ is the number of physical constraints we applied, and $\alpha$ is its associated coefficient. In our practice, coefficients are determined using \textbf{Optuna}, a hyperparameter optimisation framework, where values are randomly selected within specified ranges in each iteration to identify optimal parameters efficiently and refine model performance.

Here we introduce two physical constraints, energy conservation and flow operator.

\subsubsection{Energy Conservation}
Energy conservation is a crucial physical constraint in many applications of physical models, such as flow simulations~\cite{palm2022facilitating} and heat transfer simulations~\cite{costa2021efficient}. This principle dictates that the total energy in a system remains unchanged over time, especially in isolated scenarios where no external forces or energy transfers are present. Therefore, in a data-driven model, the constraint of energy conservation can be integrated into the loss function by defining an appropriate energy conservation regularization term~\cite{laubscher2022application}. Therefore, we define an energy conservation loss function $l_\textrm{energy}$ to measure the gap between the energy of the output data $E_\textrm{out}$ and the input data $E_\textrm{in}$, and then add this loss term with a coefficient to the total loss function as demonstrated in Eq.~\ref{eq:Loss_Funtion_with_PC}. For a single prediction step, we get:

\begin{gather}
  E_{in} = \frac{1}{k_\textrm{in}}\sum_{i=t} ^{t+k_\textrm{in}-1} \mathcal{E}(\mathcal{F}_d({\boldsymbol{\eta}}_i))  \quad \textrm{and} \quad E_\textrm{out} = \frac{1}{k_\textrm{out}}\sum_{i=t+k_\textrm{in}} ^{t+k_\textrm{in}+k_\textrm{out}-1} \mathcal{E}(\mathcal{F}_d(\tilde{\boldsymbol{\eta}}_i)) \notag \\
  l_\textrm{energy}([{\boldsymbol{\eta}}_{t:t+k_\textrm{in}-1}], [\tilde{\boldsymbol{\eta}}_{t+k_\textrm{in}:t+k_\textrm{in}+k_\textrm{out}-1}]) = \mid E_\textrm{in} - E_\textrm{out} \mid \label{eq:EC}
\end{gather}

where $\mathcal{E}$ denotes the function used to compute the total energy, consisting of both potential and kinetic energy, and $\mid \cdot \mid$ represents the absolute value.
\subsubsection{Flow Operator}
Flow operators~\cite{cai_physics-informed_2022}, denoted as ${f}$, usually appear in fluid mechanics problems, such as the shallow water equations~\cite{qi2023physics}, and the Burgers' equation. In such problems, flow operators can be used to describe the change of properties such as velocity field and pressure field of the fluid with time. In our work, we've adopted a seq2seq prediction framework that simultaneously predicts continuous time steps, simulating the temporal evolution of fluid behaviors. We anticipate that the relationships between results of multiple time steps within single output adhere to the underlying flow operators. Therefore, we apply this operator to the last element of the input sequence ${\boldsymbol{\eta}}_{t+k_\textrm{in}-1}$ (the single prediction step is demonstrated in Eq.~\ref{eq:Loss_Funtion_with_PC}), calculating the sequence output that would be derived from solving the associated PDE. The deviation between this physically-driven output and the model's prediction is then incorporated into the loss term, $l_\textrm{flow}$. Our model ensures both physical consistency and alignment of its predictions with the underlying physics described by the PDE. For a single prediction step, we get:

\begin{gather}
  \mathbf{x}_{t+k_\textrm{in}}^\textrm{fp} = f(\mathcal{F}_d({\boldsymbol{\eta}}_{t+k_\textrm{in}-1})),\quad \mathbf{x}_{t+k_\textrm{in}+1}^\textrm{fp} = f(\mathbf{x}_{t+k_\textrm{in}}^\textrm{fp}), \quad \ldots \ldots \notag \\
  l_\textrm{flow}([{\boldsymbol{\eta}}_{t:t+k_\textrm{in}-1}], [\tilde{\boldsymbol{\eta}}_{t+k_\textrm{in}:t+k_\textrm{in}+k_\textrm{out}-1}]) = \frac{1}{k_\textrm{out}}\sum_{i=t+k_\textrm{in}}^{t+k_\textrm{in}+k_\textrm{out}-1} \parallel \mathbf{x}_{i}^{fp}- \mathcal{F}_d(\boldsymbol{\eta}_{i})\parallel_2^2 \label{eq:Flow_Operator}
\end{gather}
where $\mathbf{x}^\textrm{fp}$ is the flow prediction data.

When considering physical constraints, it is necessary to decode the hidden representations back into physical space, where the physical laws are applicable, as indicated by Eq.~\ref{eq:EC} and Eq.~\ref{eq:Flow_Operator}. In this process, due to the high dimension of original data, the implement of physical constraints necessitates a substantial amount of computation resources~\cite{conti2023, liu2019sub}. If an interaction between high-fidelity and low-fidelity data were established, the physical constraints could be employed on low-fidelity physical space, which can definitely decrease the cost of utilisation of physical constraints. The establishment of such an interaction presents the potential to unlock substantial efficiency improvements in computer modelling. With this motivation, we present our innovative methodology in the subsequent parts, which aims to establish a connection between high-fidelity and low-fidelity data, while leveraging the advantages offered by each domain.

\section{Multi-Scale Physics-Constrained Neural Network}
\label{sec:MSPCNN}
\begin{figure}
    \begin{center}
        \includegraphics[width=0.8\textwidth]{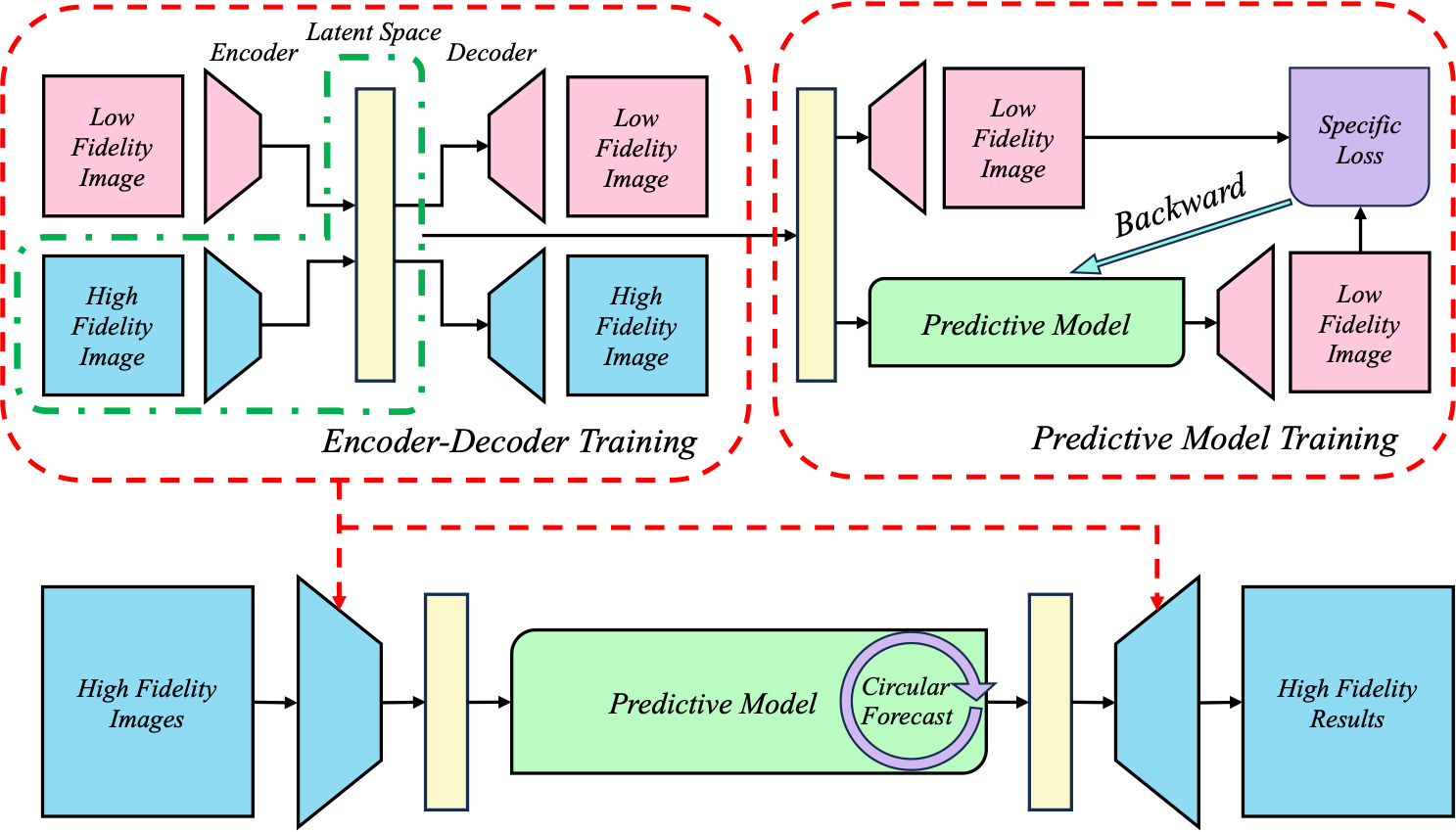}
    \end{center}
    \caption{\label{fig:MSPC_flowchart} Flowchart of MSPCNN}
\end{figure}

Now, we will introduce our newly proposed MSPCNN in detail. To clarify the main innovative design of MSPCNN, the flowchart is shown in Fig.~\ref{fig:MSPC_flowchart}. It can be seen that the main differences between the MSPCNN and PCNN are the training process of CAEs and the implementation of physical constraints.

\subsection{Multi-Fidelity CAE}
Conventional models commonly employ a CAE to handle a singular level of data fidelity. This paper presents a multi-fidelity CAE architecture, as demonstrated in Fig.~\ref{fig:MSPC_flowchart} Encoder-Decoder Training part, that comprises two separate CAEs, each specifically tailored for processing high-fidelity or low-fidelity input, respectively. The fundamental aspect of this design lies in the fact that despite the distinct levels of fidelity at which the two CAEs operate, they both facilitate the transformation of data into a latent space that is shared between them. Consequently, this shared latent space enables the identical representation between data from high- and low-fidelity fields of the same phenomenon.

Explicitly, a CAE is developed specifically for the purpose of handling high-fidelity data firstly. In this context, the encoder $\mathcal{F}_{h,e}$ is responsible for compressing the original high-fidelity data $\mathbf{x}_{h,t}$ into the latent space, resulting in the latent representation $\boldsymbol{\eta}_t$. Afterwards, the decoder $\mathcal{F}_{h,d}$ employs the latent representation to recover the initial data, resulting in \(\mathbf{x}_{h,t}^r\). In order to train the CAE, a loss function $\mathcal{J}(\theta_{\mathcal{F}_{h,e}}, \theta_{\mathcal{F}_{h,d}})$ based on MSE is employed. The objective of this loss function is to minimise the discrepancy between the reconstructed data and the original data, as seen in Eq.~\ref{eq:High_Diam_CAE}.

\begin{gather}
    \eta_t=\mathcal{F}_{h,e}(\mathbf{x}_{h,t}) \quad \textrm{and} \quad \mathbf{x}_{h,t}^r = \mathcal{F}_{h,d}(\boldsymbol{\eta}_t) \notag \\
    \mathcal{J}(\theta_{\mathcal{F}_{h,e}}, \theta_{\mathcal{F}_{h,d}}) = \frac{1}{\textrm{N}_\textrm{step}}\sum_{i=1}^{\textrm{N}_\textrm{step}} \parallel \mathit{\mathbf{x}}^r_{h,i} - \mathbf{x}_{h,i} \parallel_2^2 
    \label{eq:High_Diam_CAE}
\end{gather}

The peculiarity of the CAE for the low-fidelity data lies in its objective to align with the latent space of the CAE for the high-fidelity data. In other words, these CAEs are compressing data in different levels of fidelity into a shared latent space. In order to achieve this objective, the training process initially focuses solely on the encoder $\mathcal{F}_{l,e}$, which is responsible for compressing the low-fidelity data $\mathbf{x}_{l,t}$ into the latent space that is obtained using the high-fidelity data. The design of the loss function is characterized by its distinctiveness, as it strives to minimize the discrepancy between the low-fidelity data representation in the latent space and the corresponding representation of the high-fidelity data.

Subsequently, the decoder $\mathcal{F}_{l,d}$ is trained separately for the low-fidelity data. The objective is to restore the low-fidelity data from the shared latent space. Once again, the MSE is utilised in order to minimize the discrepancy between the reconstructed data and the original data, as demonstrated in Eq.~\ref{eq:Low_Diam_CAE}.

\begin{align}
    &\textrm{Encoder Training:} \sep &\eta_{l,t} = \mathcal{F}_{l,e}(\mathbf{x}_{l,t}) \quad &\textrm{and} \sep &\mathcal{J}(\theta_{\mathcal{F}_{l,e}}) &= \frac{1}{\textrm{N}_\textrm{step}}\sum_{i=1}^{\textrm{N}_\textrm{step}} \parallel {\boldsymbol{\eta}}_{l,i} - \boldsymbol{\eta}_{i} \parallel_2^2 \notag \\
    &\textrm{Decoder Training:} &\sep \mathbf{x}^r_{l,t} = \mathcal{F}_{l,d}(\boldsymbol{\eta}_{t}) \quad &\textrm{and} \sep &\mathcal{J}(\theta_{\mathcal{F}_{l,d}}) &= \frac{1}{\textrm{N}_\textrm{step}}\sum_{i=1}^{\textrm{N}_\textrm{step}} \parallel \mathbf{x}^r_{l,i}-\mathbf{x}_{l,i} \parallel_2^2   \label{eq:Low_Diam_CAE}
\end{align}

The algorithm of multi-fidelity CAE can be referenced as Algorithm~\ref{alg:Multi-layer CAE} in this study. In summary, the approach commences by conducting training on the first CAE using high-fidelity data. Subsequently, the encoder of second CAE is trained using low-fidelity data, while the decoder is trained using high-fidelity data, which are first encoded into the shared latent space through the high-fidelity encoder.

\begin{algorithm}
\caption{Training of Multi-Fidelity CAE in MSPCNN}
\label{alg:Multi-layer CAE}
\begin{algorithmic}[1]
\State \textbf{Inputs:}
\State High-fidelity dataset: $\mathbf{X}_{h,\textrm{train}} = [\mathbf{x}_{h,1},\mathbf{x}_{h,2},\ldots,\mathbf{x}_{h,\textrm{N}_\textrm{step}}]$
\State Low-fidelity dataset: $\mathbf{X}_{l,\textrm{train}} = [\mathbf{x}_{l,1},\mathbf{x}_{l,2},\ldots,\mathbf{x}_{l,\textrm{N}_\textrm{step}}]$

\State \textbf{Parameters:}
\State Initial learning rate: $\tau_0$
\State Epoch size: $N_{\textrm{epoch}}$
\State Initial weight parameters for encoders-decoders: $\theta_{\mathcal{F}_{h,e}}, \theta_{\mathcal{F}_{h,d}}, \theta_{\mathcal{F}_{l,e}}, \theta_{\mathcal{F}_{l,d}}$
\State \textbf{Algorithm:}

\Procedure{TrainMultiFidelityCAE}{}

    \Comment{Training High-fidelity CAE}
    \For{epoch = 1 to $N_{\textrm{epoch}}$}
        \State Compute $\boldsymbol{\eta}_t$: $\boldsymbol{\eta}_t=\mathcal{F}_{h,e}(\mathbf{x}_{h,t})$
        \State Compute $\mathbf{x}_{h,t}^r$: $\mathbf{x}_{h,t}^r = \mathcal{F}_{h,d}(\boldsymbol{\eta}_t)$
        \State Compute loss: $\mathcal{J}(\theta_{\mathcal{F}_{h,e}}, \theta_{\mathcal{F}_{h,d}}) = \frac{1}{\textrm{N}_\textrm{step}}\sum_{i=1}^{\textrm{N}_\textrm{step}} \parallel \mathit{\mathbf{x}}^r_{h,i} - \mathbf{x}_{h,i} \parallel_2^2$
        \State Update parameters $\theta_{\mathcal{F}_{h,e}}, \theta_{\mathcal{F}_{h,d}}$ using Adam optimiser
    \EndFor

    \Comment{Training Low-fidelity CAE}
    \For{epoch = 1 to $N_{\textrm{epoch}}$}
        \State Obtain $\boldsymbol{\eta}_t$ using high-fidelity encoder: $\boldsymbol{\eta}_t=\mathcal{F}_{h,e}(\mathbf{x}_{h,t})$
        \State Compute $\boldsymbol{\eta}_{l,t}$: $\boldsymbol{\eta}_{l,t} = \mathcal{F}_{l,e}(\mathbf{x}_{l,t})$
        \State Compute loss for encoder: $\mathcal{J}(\theta_{\mathcal{F}_{l,e}}) = \frac{1}{\textrm{N}_\textrm{step}}\sum_{i=1}^{\textrm{N}_\textrm{step}} \parallel \boldsymbol{\eta}_{l,j} - \boldsymbol{\eta}_{i} \parallel_2^2$
        \State Update encoder parameters $\theta_{\mathcal{F}_{l,e}}$ using Adam optimiser 
        \\
        \State Compute $\mathbf{x}^r_{l,t}$: $\mathbf{x}^r_{l,t} = \mathcal{F}_{l,d}(\boldsymbol{\eta}_{t})$
        \State Compute loss for decoder: $\mathcal{J}(\theta_{\mathcal{F}_{l,d}}) = \frac{1}{\textrm{N}_\textrm{step}}\sum_{i=1}^{\textrm{N}_\textrm{step}} \parallel \mathbf{x}^r_{l,i}-\mathbf{x}_{l,i} \parallel_2^2$
        \State Update decoder parameters $\theta_{\mathcal{F}_{l,d}}$ using Adam optimiser
    \EndFor

\EndProcedure
\end{algorithmic}
\end{algorithm}

\subsection{LSTM in the shared latent space}
The LSTM plays a pivotal role in processing the sequential data mapped into the fixed latent space by the two CAEs which have been trained at last stage, serving as the primary structure for predicting evolution. When applying physical constraints, the latent representation outputs are then decoded to low-fidelity prediction via the low-fidelity decoder. This allows for the evaluation of physical constraint errors in the low-fidelity level as shown in Eq.~\ref{eq:Low_Diam_LSTM}.

\begin{gather}
    [\tilde{\boldsymbol{\eta}}_{t+k_\textrm{in}:t+k_\textrm{in}+k_\textrm{out}-1}] = \mathcal{F}_\textrm{LSTM}([{\boldsymbol{\eta}}_{t:t+k_\textrm{in}-1}]) \notag \\
    \begin{aligned}
        \hspace{-1cm}\mathcal{J}(\theta_\textrm{LSTM}) &= l_\textrm{data}([{\boldsymbol{\eta}}_{t+k_\textrm{in}:t+k_\textrm{in}+k_\textrm{out}-1}], [\tilde{\boldsymbol{\eta}}_{t+k_\textrm{in}:t+k_\textrm{in}+k_\textrm{out}-1}]) \\
        &\hspace{1cm}+ \alpha_1  l_\textrm{energy}([{\boldsymbol{\eta}}_{t:t+k_\textrm{in}-1}], [\tilde{\boldsymbol{\eta}}_{t+k_\textrm{in}:t+k_\textrm{in}+k_\textrm{out}-1}]) \\
        &\hspace{1cm}+ \alpha_2  l_\textrm{flow}([{\boldsymbol{\eta}}_{t:t+k_\textrm{in}-1}], [\tilde{\boldsymbol{\eta}}_{t+k_\textrm{in}:t+k_\textrm{in}+k_\textrm{out}-1}])
    \end{aligned}
    \label{eq:Low_Diam_LSTM}
\end{gather}
where $\alpha_1$ and $\alpha_2$ are the associated coefficient of $l_\textrm{energy}$ and $l_\textrm{flow}$.

For the energy conservation regularization, the low-fidelity constraint is derived from Eq.~\ref{eq:EC} as manifested in Eq.~\ref{eq:Low_Diam_LSTM_energy}:

\begin{gather}
    \begin{aligned}
        l_\textrm{energy}(&[{\boldsymbol{\eta}}_{t:t+k_\textrm{in}-1}], [\tilde{\boldsymbol{\eta}}_{t+k_\textrm{in}:t+k_\textrm{in}+k_\textrm{out}-1}]) \\
        &= \mid E_\textrm{in} - E_\textrm{out} \mid \\
        &= \left| \frac{1}{k_\textrm{in}}\sum_{i=t} ^{t+k_\textrm{in}-1} \mathcal{E}(\mathcal{F}_{l,d}({\boldsymbol{\eta}}_i))  - \frac{1}{k_\textrm{out}}\sum_{i=t+k_\textrm{in}} ^{t+k_\textrm{in}+k_\textrm{out}-1} \mathcal{E}(\mathcal{F}_{l,d}(\tilde{\boldsymbol{\eta}}_i)) \right|
    \end{aligned}
    \label{eq:Low_Diam_LSTM_energy}
\end{gather}

Furthermore, for the flow operation regularization, the low-fidelity constraint can be get from Eq.~\ref{eq:Flow_Operator} as manifested in Eq.~\ref{eq:Low_Diam_LSTM_flow_operator}:

\begin{gather}
    \mathbf{x}_{l,t+k_\textrm{in}}^\textrm{fp} = f_l(\mathcal{F}_d({\boldsymbol{\eta}}_{t+k_\textrm{in}-1})),\quad \mathbf{x}_{l,t+k_\textrm{in}+1}^\textrm{fp} = f_l(\mathbf{x}_{l,t+k_\textrm{in}}^{fp}), \quad \ldots \ldots \notag \\
    l_\textrm{flow}([{\boldsymbol{\eta}}_{t:t+k_\textrm{in}-1}] [\tilde{\boldsymbol{\eta}}_{t+k_\textrm{in}:t+k_\textrm{in}+k_\textrm{out}-1}]) = \frac{1}{k_\textrm{out}}\sum_{i=t+k_\textrm{in}}^{t+k_\textrm{in}+k_\textrm{out}-1} \parallel \mathbf{x}_{l,i}^\textrm{fp} - \mathcal{F}_{l,d}(\boldsymbol{\eta}_{i}) \parallel_2^2
    \label{eq:Low_Diam_LSTM_flow_operator}
\end{gather}
where $f_l$ represents the flow operator in low-fidelity field, $\mathbf{x}_{l}^\textrm{fp}$ is the flow prediction data in low-fidelity field. The process of the algorithm of LSTM is summarised in Algorithm~\ref{alg:Training Seq2Seq LSTM}.

Additionally, the output from the predictive model (LSTM) remains in the form of latent representations. To obtain the final predictions in the full physical space, these representations must be passed through a decoder, as illustrated in Fig.~\ref{fig:MSPC_flowchart} to gain the final outputs. The specific loss of the LSTM can overall be written as:

\begin{align}
    \mathcal{J}(\theta_\textrm{LSTM}) &= l_\textrm{data}([{\boldsymbol{\eta}}_{t+k_\textrm{in}:t+k_\textrm{in}+k_\textrm{out}-1}], [\tilde{\boldsymbol{\eta}}_{t+k_\textrm{in}:t+k_\textrm{in}+k_\textrm{out}-1}]) \notag \\
    &\hspace{1cm}+ \alpha_1  l_\textrm{energy}([{\boldsymbol{\eta}}_{t:t+k_\textrm{in}-1}], [\tilde{\boldsymbol{\eta}}_{t+k_\textrm{in}:t+k_\textrm{in}+k_\textrm{out}-1}])\notag \\
    &\hspace{1cm}+ \alpha_2  l_\textrm{flow}([{\boldsymbol{\eta}}_{t:t+k_\textrm{in}-1}], [\tilde{\boldsymbol{\eta}}_{t+k_\textrm{in}:t+k_\textrm{in}+k_\textrm{out}-1}])\notag \\
    &= \frac{1}{k_\textrm{out}} \sum_{i=t+k_\textrm{in}}^{t+k_\textrm{in}+k_\textrm{out}-1} \lVert {\boldsymbol{\eta}}_i - \tilde{\boldsymbol{\eta}}_i \rVert_2^2 \notag \\
    &\hspace{1cm}+ \alpha_1 \left| \frac{1}{k_\textrm{in}}\sum_{i=t} ^{t+k_\textrm{in}-1} \mathcal{E}(\mathcal{F}_{l,d}({\boldsymbol{\eta}}_i))  - \frac{1}{k_\textrm{out}}\sum_{i=t+k_\textrm{in}} ^{t+k_\textrm{in}+k_\textrm{out}-1} \mathcal{E}(\mathcal{F}_{l,d}(\tilde{\boldsymbol{\eta}}_i)) \right| \notag \\
    &\hspace{1cm}+ \alpha_2 \frac{1}{k_\textrm{out}}\sum_{i=t+k_\textrm{in}}^{t+k_\textrm{in}+k_\textrm{out}-1} \parallel \mathbf{x}_{l,i}^\textrm{fp} - \mathcal{F}_{l,d}(\boldsymbol{\eta}_{i}) \parallel_2^2
\label{specific_loss_LSTM}
\end{align}

\begin{algorithm}
\caption{Training of Seq2Seq LSTM in MSPCNN}
\label{alg:Training Seq2Seq LSTM}
\begin{algorithmic}[1]
\State \textbf{Inputs:}
\State High-fidelity training sequence data: $\mathbf{X}_{h,\textrm{train}}$
\State Fixed Encoder for high-fidelity: $\mathcal{F}_{h,e}$
\State Fixed Decoder for low-fidelity: $\mathcal{F}_{l,d}$
\State \textbf{Parameters:}
\State Number of physical constraints: c
\State Physical constraints: $l_{\textrm{data}}, l_{\textrm{energy}}, l_{\textrm{flow}}$
\State Weights for physical constraints: $\alpha_1, \alpha_2$
\State Initial learning rate: $\tau_0$
\State Epoch size: $N_{\textrm{epoch}}$
\State Sequence input length: $k_{\textrm{in}}$
\State Sequence output length: $k_{\textrm{out}}$
\State Initial weight parameters for LSTM: $\theta_{\textrm{LSTM}}$
\State \textbf{Algorithm:}

\Procedure{TrainSeq2SeqLSTM}{}
    \For{epoch = 1 to $N_{\textrm{epoch}}$}
        \For{$t$ in 1 to length($\mathbf{X}_{l,\textrm{train}}$) - $k_{\textrm{in}}$ - $k_{\textrm{out}}$ + 1}
            \State Extract sequence from input: $\mathbf{x}_{h,t:t+k_{\textrm{in}}-1}$ and target $\mathbf{x}_{h,t+k_{\textrm{in}}:t+k_{\textrm{in}}+k_{\textrm{out}}-1}$
            \State Convert high-fidelity input to latent space: $\boldsymbol{\eta}_{t:t+k_{\textrm{in}}-1} = \mathcal{F}_{h,e}(\mathbf{x}_{h,t:t+k_{\textrm{in}}-1})$
            \State Compute LSTM output: $ \boldsymbol{\tilde \eta}_{t+k_{\textrm{in}}:t+k_{\textrm{in}}+k_{\textrm{out}}-1} = \textrm{LSTM}(\boldsymbol{\eta}_{t:t+k_{\textrm{in}}-1}; \theta_{\textrm{LSTM}})$
            \State Convert LSTM output to low-fidelity: 
            \Statex \hspace{10em}$\mathbf{x}^r_{l,t+k_{\textrm{in}}:t+k_{\textrm{in}}+k_{\textrm{out}}-1} = \mathcal{F}_{l,d}(\boldsymbol{\tilde \eta}_{t+k_{\textrm{in}}:t+k_{\textrm{in}}+k_{\textrm{out}}-1})$
            \State Compute loss: 
            \Statex \hspace{10em} $\mathcal{J} = l_\textrm{data}([{\boldsymbol{\eta}}_{t+k_\textrm{in}:t+k_\textrm{in}+k_\textrm{out}-1}], [\tilde{\boldsymbol{\eta}}_{t+k_\textrm{in}:t+k_\textrm{in}+k_\textrm{out}-1}])$
            \Statex \hspace{12em} $+ \sum_j^c \alpha_j l_\textrm{physics}^j([{\boldsymbol{\eta}}_{t:t+k_\textrm{in}-1}], [\tilde{\boldsymbol{\eta}}_{t+k_\textrm{in}:t+k_\textrm{in}+k_\textrm{out}-1}])$
            \Statex \hspace{12em} $= \frac{1}{k_\textrm{out}} \sum_{i=t+k_\textrm{in}}^{t+k_\textrm{in}+k_\textrm{out}-1} \lVert {\boldsymbol{\eta}}_i - \tilde{\boldsymbol{\eta}}_i \rVert_2^2$ 
            \Statex \hspace{12em} $+ \alpha_1 \mathit{l_{\textrm{energy}}} + \alpha_2 \mathit{l_{\textrm{flow}}} $

            \State Update LSTM parameters $\theta_{\textrm{LSTM}}$ using Adam optimiser
        \EndFor
    \EndFor
\EndProcedure

\end{algorithmic}
\end{algorithm}

Overall, compared with the PCNN, central to our proposed method is the strategic use of a shared latent space achieved by leveraging multi-fidelity CAE. This shared latent space is essential as it facilitates the smooth mapping of data across different fidelities. In other words, various fidelities data can get the same latent representation with different encoders, and the compressed data can also be decoded into either a low-fidelity or high-fidelity space as desired. With such a characteristic, predictive model can leverage both high- and low-fidelity data for training simultaneously and the physical constraints can be applied in low-fidelity level for high-fidelity surrogate model training. By applying physical constraints at the low-fidelity level, significant training costs can be saved compared to imposing them at the high-fidelity level. Furthermore, MSPCNN maintains the LSTM's structure intact throughout the optimisation process, ensuring that the online prediction phase remains computationally efficient and aligned with the conventional predictive models in terms of resource usage.

\section{Numerical example: Burgers' Equation}
\label{sec:burgers}
Burgers' equation is a fundamental PDE occurring in various areas, such as fluid mechanics, nonlinear acoustics, and gas dynamics. The numerical results for the Burgers' system in this paper are derived by solving the equations using spatial discretisation with backward and central difference schemes for convection and diffusion terms, respectively, and time integration using the Euler method. In our evaluation of the MSPCNN, we employ high-fidelity and low-fidelity simulations of the 2D Burgers' equation problem. Both simulations, albeit at different resolutions, depict the same physical phenomenon, with time appropriately scaled for consistency. The domain for the high-fidelity simulation is defined as a 129×129 grid, while it is 33×33 for the low-fidelity simulation. The boundaries of these squares are configured with Dirichlet boundary conditions. The viscosity is 0.01$N \cdot s \cdot m^{-2}$ and the initial velocity ranges from 1.5$m\cdot s^{-1}$ to 5$m\cdot s^{-1}$. The equations are presented as:

\begin{gather}
\label{eq:burgers}
    \frac{\partial u}{\partial t}+u\frac{\partial u}{\partial x} +v\frac{\partial u}{\partial y} = \frac{1}{Re}(\frac{\partial^2 u}{\partial x^2} + \frac{\partial^2 u}{\partial y^2}) \notag \\
    \frac{\partial v}{\partial t}+u\frac{\partial v}{\partial x} +v\frac{\partial v}{\partial y} = \frac{1}{Re}(\frac{\partial^2 v}{\partial x^2} + \frac{\partial^2 v}{\partial y^2})
\end{gather}
where $u$ and $v$ represent the velocity components and $t$ is time, $x$ and $y$ represent the coordinate system. $Re$ is the Reynolds number, which can be calculated by $Re = \frac{VL}{\upsilon}$, where $V$ is the flow speed, specified as initial velocity, $L$ is characteristic linear dimension and $\upsilon$ is viscosity.

Specifically, we use the recurrent prediction method, as shown in Eq.~\ref{eq:LSTM_Predict}, where ${k_\textrm{in}}={k_\textrm{out}}=3$, to predict Burgers' Equation. In order to deeply explore the model's performance and the impact of various constraints, we designed the following sets of controlled experiments:

\begin{enumerate}
    \item Training on high-fidelity data versus multi-fidelity data:
    \begin{enumerate}
        \item Basic LSTM (without physical constraints): Trained using pure high-fidelity data.
        \item Multi-fidelity Basic LSTM training: In order to verify whether low-fidelity data and high-fidelity data can train the model simultaneously, we use multi-fidelity data (both the high- and low-fidelity data) as the training dataset.
    \end{enumerate}
    \item Effects of a single physical constraint on the model:
    \begin{enumerate}
        \item High-fidelity constraint: We use a single physical constraint, such as energy conservation (EC) or flow operator(FO), and only apply them on high-fidelity field to explore its effect.
        \item Low-fidelity constraint: Under the same physical constraints, we apply the physical constraint in low-fidelity field to constrain high-fidelity surrogate models.
    \end{enumerate}
    \item Effect of multiple physical constraints:
    \begin{enumerate}
        \item High-Fidelity Multiple Constraints: We use multiple physical constraints, including energy conservation (EC) and flow operator(FO), and apply them on high-fidelity field to explore its effect.
        \item Low-Fidelity Multiple Constraints: Under the same physical constraints, we apply the multiple physical constraints in low-fidelity field to constrain high-fidelity surrogate models, and compare the effect with multiple physical constraints in high-fidelity field.
    \end{enumerate}
\end{enumerate}

These experiments aim to gain insight into the role and performance of low-fidelity data in model training and constraints.
\subsection{Validation of Multi-Fidelity CAE in Burgers' Equation}
\begin{figure}
    \begin{center}
        \includegraphics[width=0.8\textwidth]{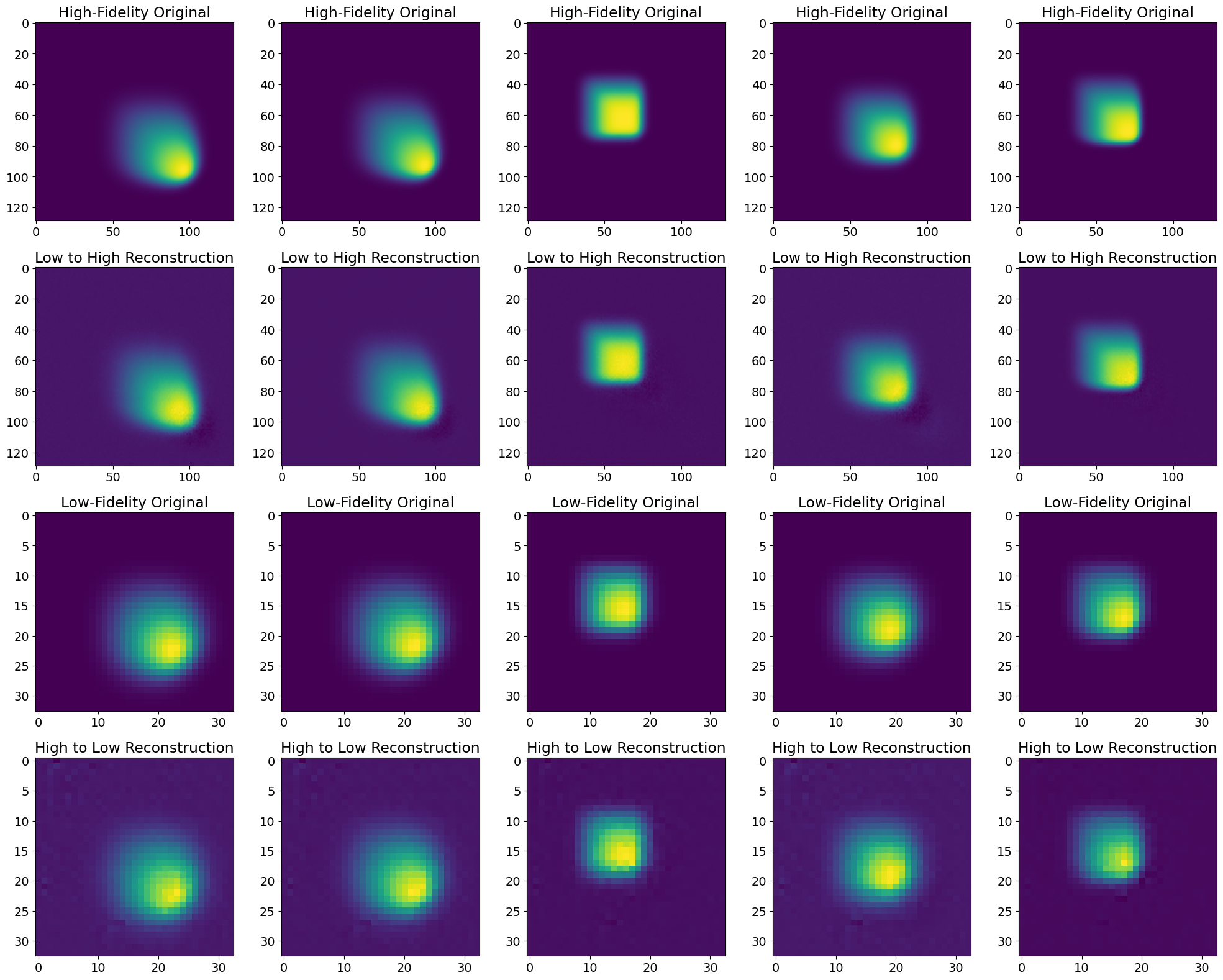}
    \end{center}
    \caption{\label{fig:BurgerCAEPrediction} Results from Multi-Fidelity CAE in Burgers' Equation ($u$ dimension)}
\end{figure}

Firstly, We showcase the efficacy of our multi-fidelity CAE in efficiently handling both high-fidelity and low-fidelity Burgers' equation data. Fig.~\ref{fig:BurgerCAEPrediction} underscores the adeptness of our multi-fidelity CAE in transforming data between various fidelity levels. The first two rows in Fig.~\ref{fig:BurgerCAEPrediction} illustrate a comparison between the original high-fidelity data and its reconstructed version derived from low-fidelity data. Similarly, the third and fourth rows display the original low-fidelity data alongside its reconstructed version obtained from high-fidelity data. The reconstructions exhibit high precision, laying a solid foundation for subsequent utilisation. These findings demonstrate that the shared latent space is capable of capturing high-fidelity field details while encoding low-fidelity data.
\subsection{Training on high-fidelity data versus multi-fidelity data}
\label{subsec:muldata_burger}
\begin{figure}
    \begin{center}
        \includegraphics[width=0.8\textwidth]{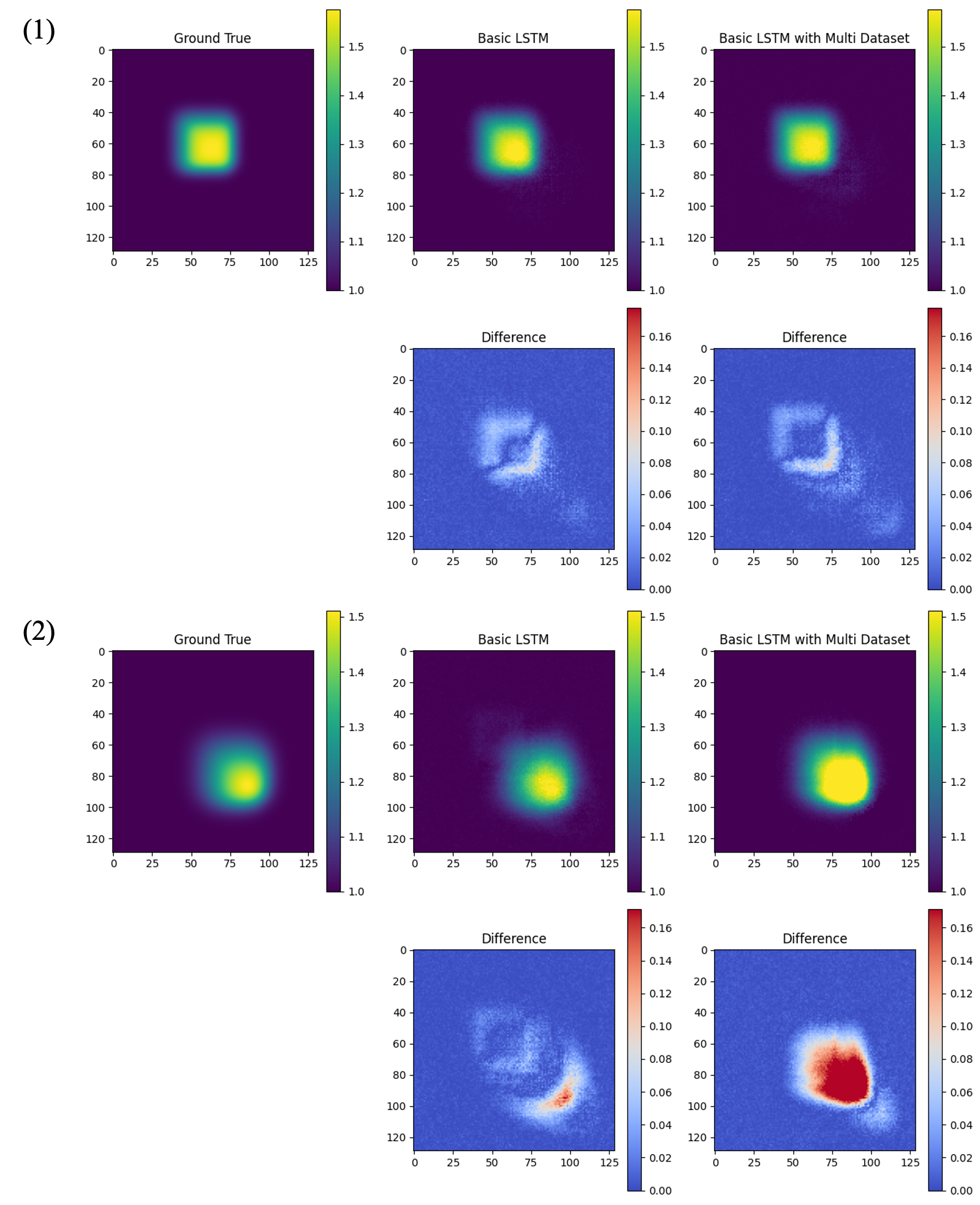}
    \end{center}
    \caption{\label{fig:basic_muldata_prediction} Prediction Results ($u$ dimension) and Difference with Groundtrue of LSTM with Multi-Fidelity Data for (1)t=25 and (2)t=99 in Burgers' System}
\end{figure}

\begin{figure}
    \begin{center}
        \includegraphics[width=0.8\textwidth]{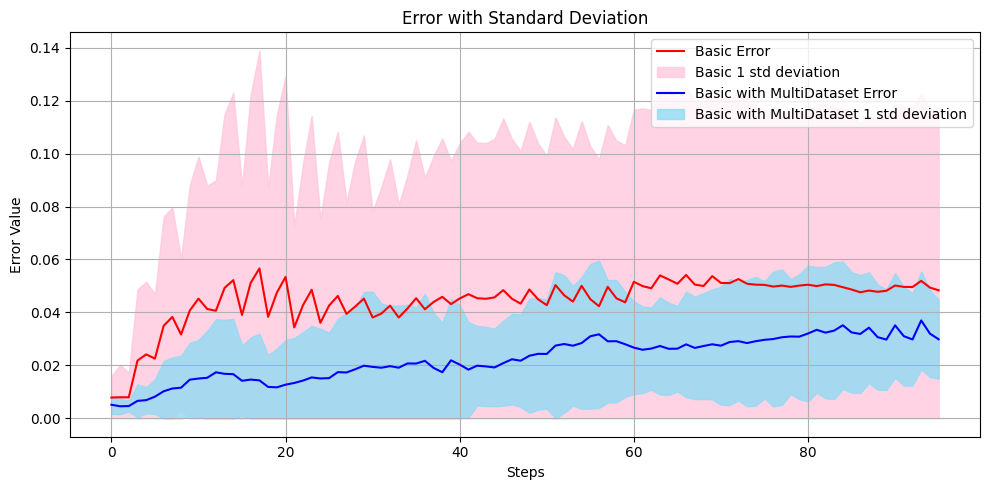}
    \end{center}
    \caption{\label{fig:basic_muldata_error} Performance of Basic LSTM with Multi-Fidelity Data Compared with Basic LSTM in Burgers' System}
\end{figure}

As illustrated in Fig.~\ref{fig:basic_muldata_prediction}, we compare a pure LSTM model using 300 high-fidelity samples against one trained with an additional 300 low-fidelity samples using the multi-scale encoder and decoder as explained in section~\ref{sec:MSPCNN}. The difference in Fig.~\ref{fig:basic_muldata_prediction} is calculated at each point as the absolute value of direct subtraction of the predicted value from the actual value, which represents the absolute error at each point. Turning to Fig.~\ref{fig:basic_muldata_error}, this graph details how the MSE and standard deviation change cumulatively as the time step increases. From Fig.~\ref{fig:basic_muldata_error}, we can clearly see that the supplement of low-fidelity data can bring a significant improvement in prediction accuracy while reducing the uncertainties represented by the transparent zones. It's important to note that our model employs a seq2seq approach for computations, meaning the output is a sequence. However, when calculating loss and standard deviation (std), we disaggregate this sequence, comparing each time step individually with the ground truth. For the loss and std, we compute the mean squared error for each predicted timestep and then calculate the std across all cycles, reflecting model performance variability over time. This method is consistently applied across all performance figures and encompasses the entire test dataset. However, in light of the statistical results, the predictions in Fig.~\ref{fig:basic_muldata_prediction} show an opposite error graph. Our observations suggest that utilising multiple datasets centralises the errors, which results in the amplification of the error peak. This phenomenon will be further analysed in subsequent sections.

\subsection{Effects of a Single Physical Constraint on the Model}

\begin{figure}
    \begin{center}
        \includegraphics[width=0.8\textwidth]{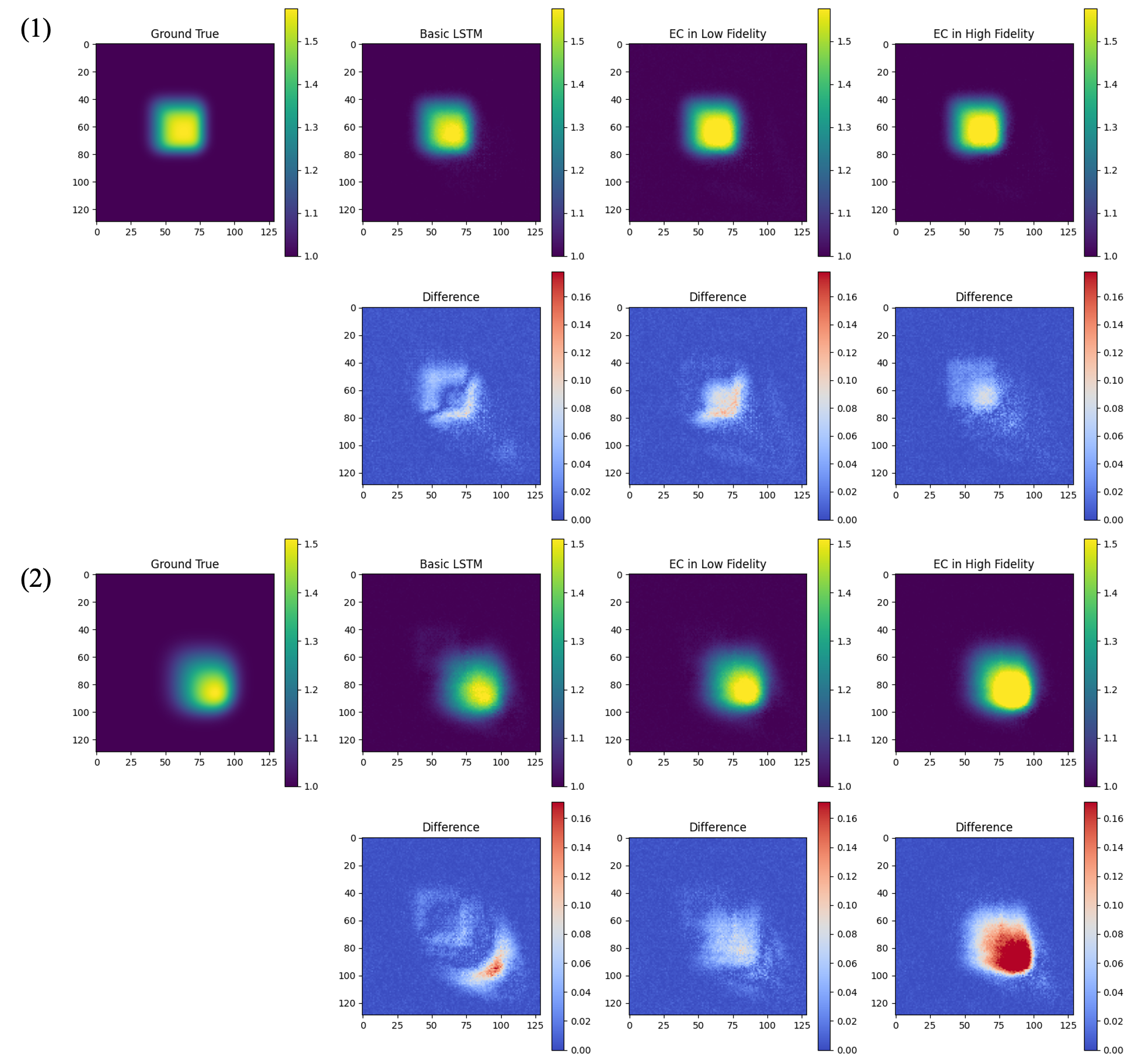}
    \end{center}
    \caption{\label{fig:burger_ec}  Prediction Results ($u$ dimension) and Difference with Groundtrue of LSTM with EC Constraint for (1)t=25 and (2)t=99 in Burgers' System}
\end{figure}

\begin{figure}
    \begin{center}
        \includegraphics[width=0.8\textwidth]{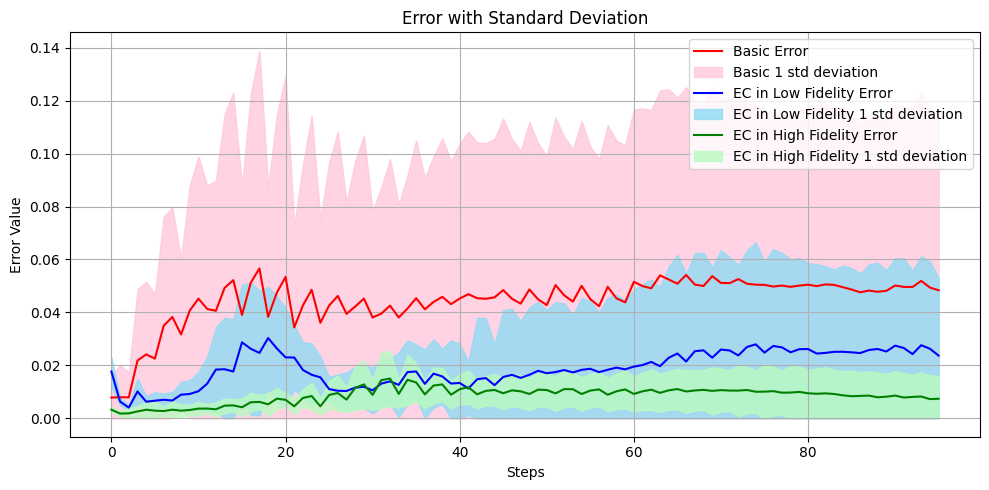}
    \end{center}
    \caption{\label{fig:ec_error} Performance of MSPCNN with EC Constraint Compared with Basic Predictive model in Burgers' System}
\end{figure}

In Fig.~\ref{fig:burger_ec}, we showcase the predictions of the MSPCNN and PCNN with the energy conservation constraint employing in low-fidelity (LF-EC) and high-fidelity (HF-EC) fields, respectively, compared with the basic LSTM and highlight the difference with the groundtruth. Furthermore, Fig.~\ref{fig:ec_error} shows the performance of these three different models in long-time prediction. Compared to the basic LSTM approach, these results show that both HF-EC and LF-EC can significantly reduce the MSE and the range of standard deviations which are visibly evident from the shaded part in Fig.~\ref{fig:ec_error}, underscoring that physical constraints not only diminish prediction error but also augment the model's robustness when applied in the training process. Referring to Table~\ref{tab:Performance Comparison between Models Burgers}, when applying the energy conservation constraint in the high-fidelity field, the MSE is reduced by nearly 85\% compared to the basic LSTM model, where the low-fidelity model demonstrates an improvement of 52\% relative to the basic model. However, by leveraging the energy conservation constraint in the low-fidelity field, our model can achieve around 60\% of high-fidelity model's performance with only 50\% of its training time.

\begin{figure}
    \begin{center}
        \includegraphics[width=0.8\textwidth]{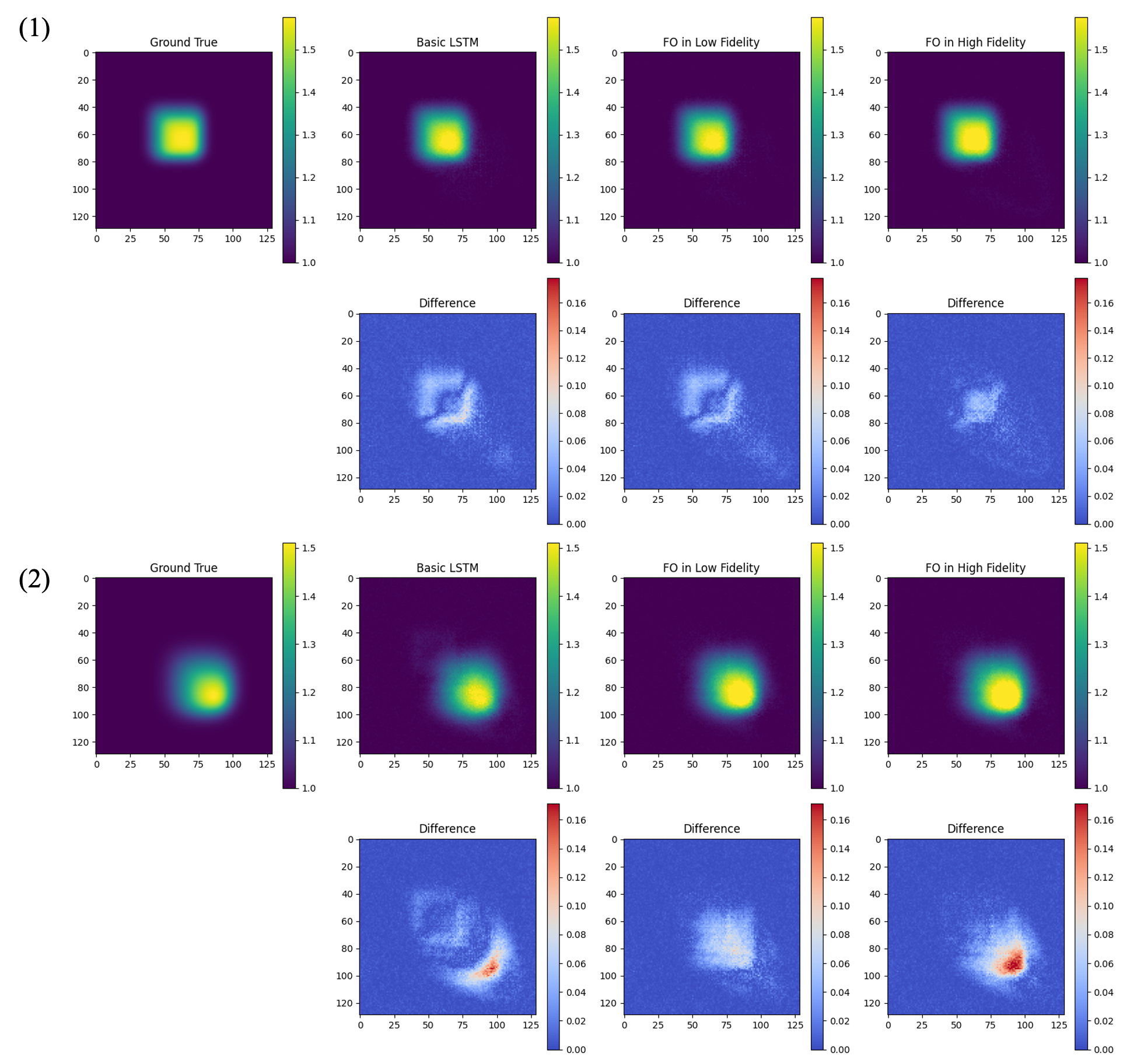}
    \end{center}
    \caption{\label{fig:burger_fo}  Prediction Results ($u$ dimension) and Difference with Groundtrue of LSTM with FO Constraint for (1)t=25 and (2)t=99 in Burgers' System}
\end{figure}

\begin{figure}
    \begin{center}
        \includegraphics[width=0.8\textwidth]{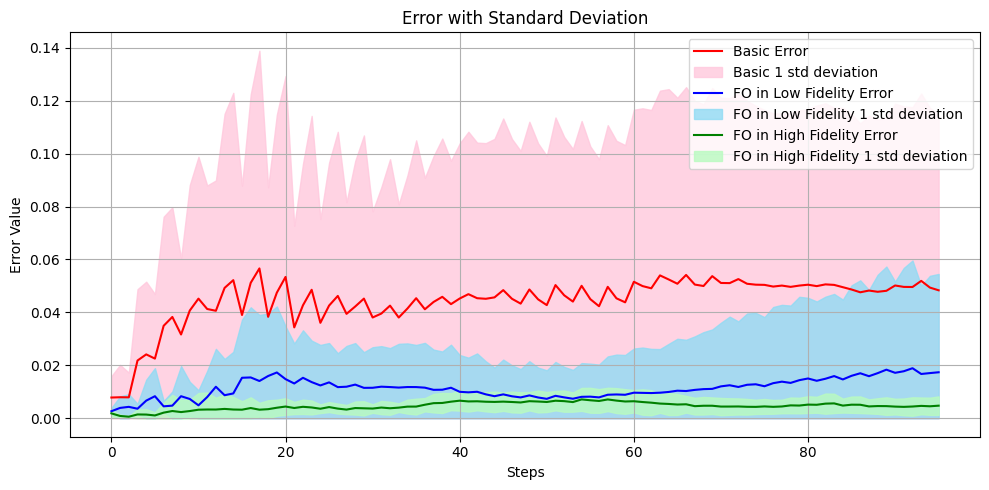}
    \end{center}
    \caption{\label{fig:fo_error} Performance of MSPCNN with FO Constraint Compared with Basic Predictive model in Burgers' System}
\end{figure}

Transitioning to Fig.~\ref{fig:burger_fo}, the prediction performances of the MSPCNN and PCNN under the constraints of low-fidelity (LF-FO) and high-fidelity (HF-FO) flow operators and their deviations from groundtrue are showcased, respectively. Fig.~\ref{fig:fo_error} and Table~\ref{tab:Performance Comparison between Models Burgers} complement the description of the cumulative trend of performance metrics and training time. Upon implementing the flow operator constraint, the MSE for LF-FO is reduced by approximately 66\% compared with the basic LSTM. Meanwhile, for HF-FO, the MSE sees a more substantial reduction, decreasing the error by over 90\%. Just as solely applying the energy conservation constraint, the shaded portion of Fig.~\ref{fig:fo_error} elucidates the range of standard deviations, reiterating the enhanced stability introduced by the physical constraints, where both HF-Fo and LF-FO outperform the basic LSTM. Remarkably, upon implementing the flow operator constraint, the low-fidelity model achieves 73\% of the high-fidelity performance while only requiring 25\% of the training time.

\begin{figure}
    \begin{center}
        \includegraphics[width=0.6\textwidth]{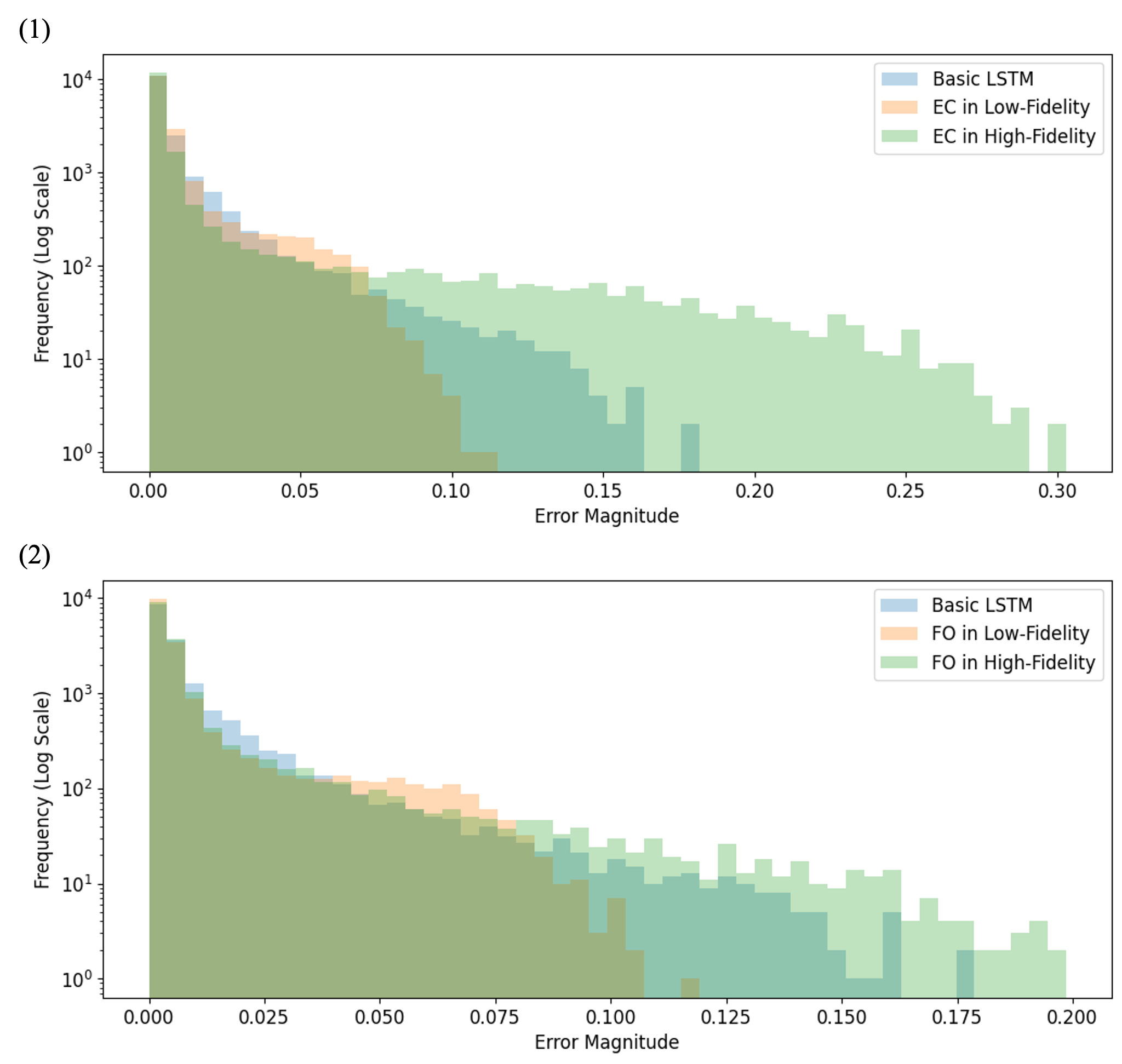}
    \end{center}
    \caption{\label{fig:ec_fo_hist} Error histogram comparison in Burgers' System (1) Energy conservation constraint: comparison of high-fidelity, low-fidelity and basic model error histograms (2) Flow operator constraint: comparison of high-fidelity, low-fidelity and basic model error histograms}
\end{figure}

It is worth noting that, comparing Fig.~\ref{fig:burger_ec} and Fig.~\ref{fig:burger_fo} with Fig.~\ref{fig:ec_error} and Fig.~\ref{fig:fo_error}, the predictions under high-fidelity physical constraints demonstrate a higher error peak, despite a lower overall MSE, which also appears in section~\ref{subsec:muldata_burger}. To further clarify this point, we plot the histogram of prediction errors for the last step of the recurrent prediction (as shown in Fig.~\ref{fig:ec_fo_hist}). From Fig.~\ref{fig:ec_fo_hist}, we observe that while the upper bound of the error (i.e., the maximum error) does increase when high-fidelity data is introduced, the frequency of low errors increases accordingly, leading to a reduction in the overall MSE. In contrast, the low-fidelity restriction strategy demonstrates superior performance in this aspect. As illustrated in Fig.~\ref{fig:ec_fo_hist}, applying physical constraints in the low-fidelity field by MSPCNN not only improves the proportion of lower errors but also doesn't result in the amplification of the error peak. Compared to the basic LSTM model, the introduction of both the energy conservation constraint and the flow operator constraint in the low-fidelity field has successfully lowered the upper bound of errors from 0.175 to around 0.11. Furthermore, the histogram reveals that the distribution quantity within the 0-0.1 range is greater than that of the basic model.

The amplification of the error peak in Burgers' equation can be attributed to several factors. While the equation describes a relatively simple process, during backpropagation, the model tends to prioritize the surrounding regions of the Burgers' system due to their similar physical characteristics. This dominance causes the model to overly focus on surrounding regions, often neglecting the central evolution area and leading to increased errors there. For example, when the flow operator is used as a physical constraint, although the error in the central evolution area increases, a certain degree of error will not have a large impact on the evolution of the entire area because the velocity of the area itself is relatively large. Nevertheless, in surrounding regions characterised by consistently low and stable velocities, the presence of a substantial error has the potential to initiate a propagating disturbance. This phenomenon has the potential to cause substantial deviation from the groundtruth across the whole surrounding region. This means that the backpropagation of the model is more accurate in these surrounding regions, resulting in lower errors in these regions, while the error increases in the central regions. Additionally, as seen in Table~\ref{tab:Performance Comparison between Models Burgers}, and Fig.~\ref{fig:basic_muldata_prediction},~\ref{fig:burger_ec},~\ref{fig:burger_fo}, this phenomenon is alleviated as the error decreases. Hence, when the error diminishes, the accuracy of predictions in the central region improves.


\begin{table}[h]
\centering
\caption{Performance Comparison between Models in Burgers' System}
\label{tab:Performance Comparison between Models Burgers}
\begin{tabular}{|c|c|c|c|c|}
\hline
Case & Model & MSE & SSIM & Training Time/Epoch (s) \\
\hline
\multirow{9}{*}{Burgers' System} & Basic & 100\% & 0.9925 & 5.97 \\
\cline{2-5}
                        & MultiDataset & 34.1\%& 0.9933 & 11.26\\
\cline{2-5}
                        & HF-EC & 15.2\%& 0.9981 & 109.45\\
\cline{2-5}
                        & LF-EC & 51.9\%& 0.9958 & 52.18\\
\cline{2-5}
                        & HF-FO & 9.4\%& 0.9988 & 60.72\\
\cline{2-5}
                        & LF-FO & 34.3\%& 0.9978 & 12.29\\
\cline{2-5}
                        & HF-MulCons & 5.2\%& 0.9989 & 164.36\\
\cline{2-5}
                        & LF-MulCons & 22.0\%& 0.9972 & 54.96\\
\hline
\end{tabular}

\smallskip 
\small{
\begin{tabular}{l}
\textbf{Note:}\\
\textbullet~ Basic: Predictive model trained by solely high-fidelity data.\\
\textbullet~ MultiDataset: Predictive model trained by both high and low-fidelity data.\\
\textbullet~ HF-EC, LF-EC: Model with energy conservation constraint in high and low-fidelity field.\\
\textbullet~ HF-FO, LF-FO: Model with flow operator constraint in high and low-fidelity field.\\
\textbullet~ HF-MulCons, LF-MulCons: Model with multiple constraints in high and low-fidelity field.\\
\textbullet~ MSE: Mean Squared Error with reference to the basic model set at 100\%.\\
\textbullet~ SSIM: Structural Similarity Index (with data range of 1.0).\\
\textbullet~ Training Time/Epoch (s): Time taken to run one epoch during training, unit: seconds. \\
\textbullet~ \parbox[t]{5in}{The coefficient of the physical constraint $\alpha$ is optimised using the validation set to achieve the best performance for each model. $\alpha_\textrm{EC}$ is the coefficient of energy conservation constraint. $\alpha_\textrm{FO}$ is the coefficient of flow operator constraint. Specifically, $\alpha_\textrm{EC}=2.0e-6$ for HF-EC, $\alpha_\textrm{EC}=2.8e-4$ for LF-EC, $\alpha_\textrm{EC}=4.3e-6$ for HF-MulCons, $\alpha_\textrm{EC}=1.1e-4$ for LF-MulCons, $\alpha_\textrm{FO}=2.5e-3$ for HF-FO, $\alpha_\textrm{FO}=8.5e-4$ for LF-FO, $\alpha_\textrm{FO}=1.2e-3$ for HF-MulCons, $\alpha_\textrm{FO}=5.0e-4$ for LF-MulCons.}

\end{tabular}
}

\end{table}
\subsection{Effect of Multiple Physical Constraints}
\begin{figure}
    \begin{center}
        \includegraphics[width=0.8\textwidth]{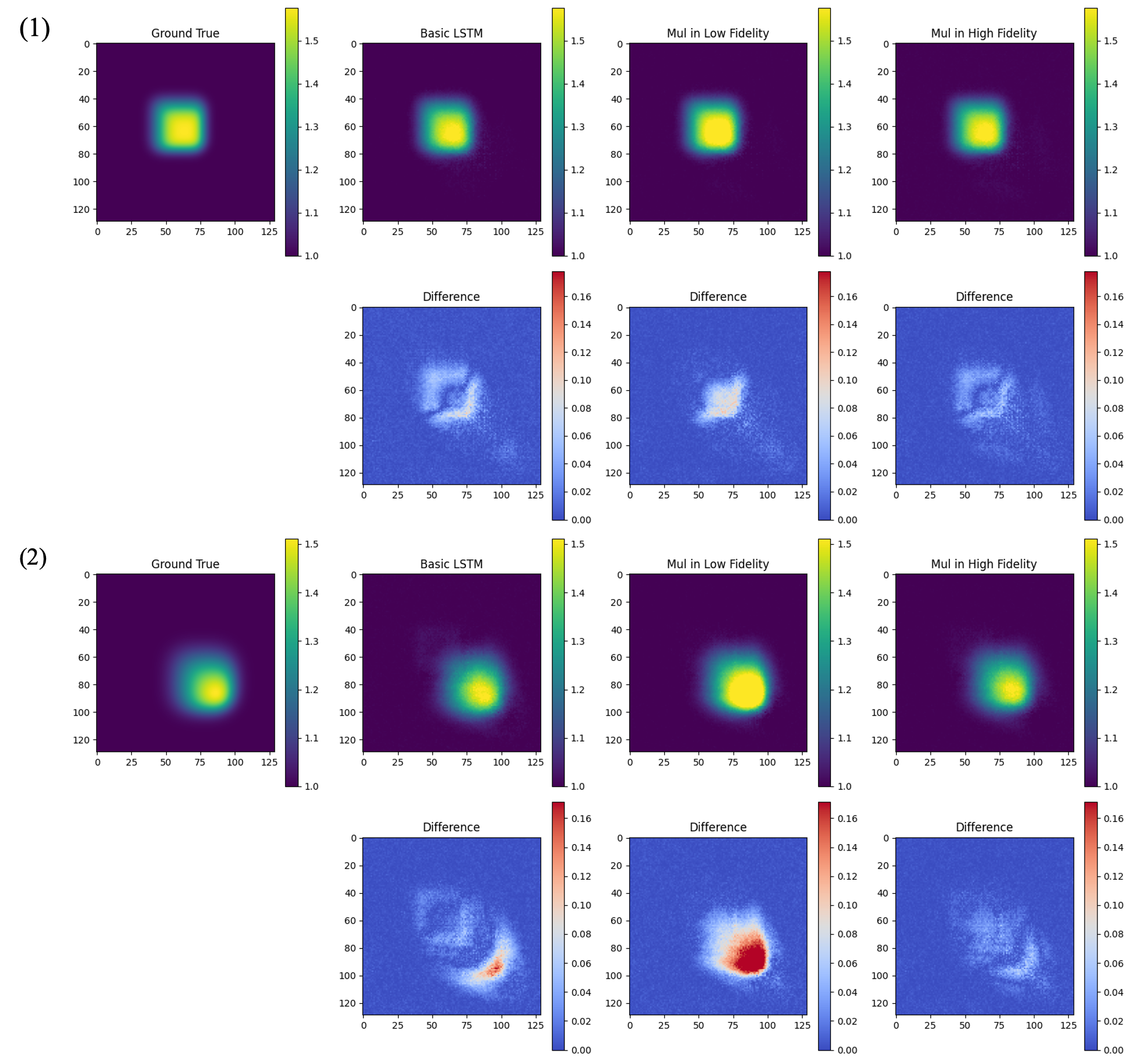}
    \end{center}
    \caption{\label{fig:burger_mul} Prediction Results ($u$ dimension) and Difference with Groundtrue of LSTM with Multiple Constraints for (1)t=25 and (2)t=99 in Burgers' System}
\end{figure}

\begin{figure}
    \begin{center}
        \includegraphics[width=0.8\textwidth]{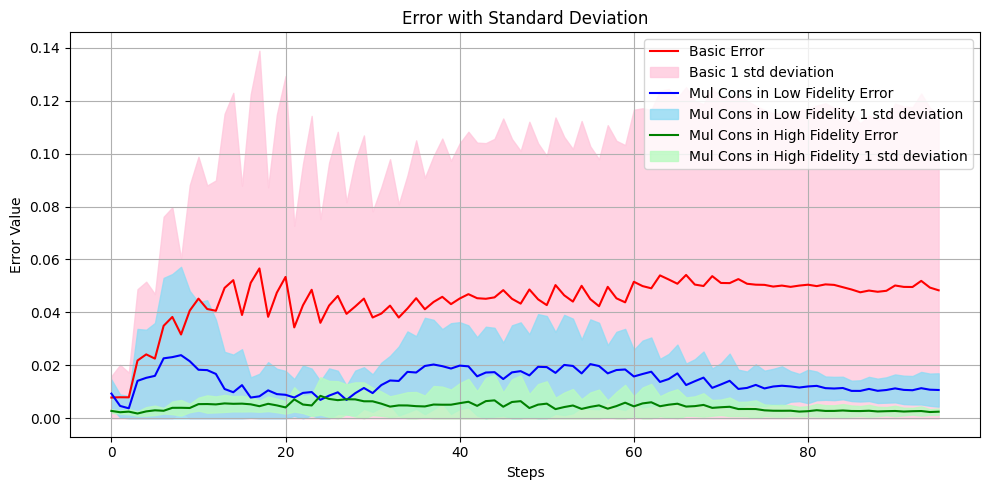}
    \end{center}
    \caption{\label{fig:multi_error} Performance of MSPCNN with Multiple Constraints Compared with Basic Predictive model in Burgers' System}
\end{figure}

The application of a single physical constraint has been shown to improve the long-time predictive accuracy of the model. To explore the effect of applying multiple physical constraints, we further build models incorporating both "energy conservation" and "flow operator" constraints, and test them in the high- (HF-MulCons) and low-fidelity (LF-MulCons) field. The corresponding prediction results are shown in Fig.~\ref{fig:burger_mul}, while Fig.~\ref{fig:multi_error} details the cumulative change of MSE and standard deviation with increasing time steps. Notably, when comparing Fig.~\ref{fig:multi_error} with Fig.~\ref{fig:ec_error},~\ref{fig:fo_error}, it becomes evident that upon employing multiple physical constraints, the disparity between the low-fidelity model and the high-fidelity model is markedly reduced compared to scenarios with a single physical constraint. Table~\ref{tab:Performance Comparison between Models Burgers} demonstrates more statistical details. The LF-MulCons model reaches about 80\% of the HF-MulCons model's accuracy while requiring only 33.5\% of its training time per epoch. When comparing LF-MulCons to LF-EC and LF-FO, it is shown that LF-MulCons delivers superior MSE performance, with a slight rise in computational requirements. This finding shows that our model is capable of providing a compromise between accuracy and computational demand in multi-constraint scenarios.

Interestingly, in Fig.~\ref{fig:burger_mul}, the amplification of the error peak previously observed in high-fidelity single-physics-confined PCNN now appears in low-fidelity multiple-physics-confined MSPCNN. It is noteworthy that by employing the low-fidelity multiple-physics-confined MSPCNN, we can achieve the predictive performance of the high-fidelity single-physics-confined PCNN. This suggests that multiple constraints at a low-fidelity field can potentially substitute for a single or fewer constraints at a high-fidelity field. Moreover, with further improvement in prediction accuracy, the high-fidelity multiphysics-constrained PCNN successfully addresses the amplification of the error peak. This observed behaviour aligns with and validates our hypothesis regarding the model's tendencies during backpropagation in the context of the Burgers' equation. 

Overall, MSPCNN showcases its ability to integrate data across different fidelities to train a high-fidelity predictive model, thereby enhancing its accuracy. Furthermore, when implementing MSPCNN with low-fidelity physical constraints in the Burgers' system, it becomes evident that the model effectively strikes a balance between accuracy and computational efficiency.


\section{Numerical example: Shallow Water}
\label{sec:sw}
In the previous section, we showed that MSPCNN can efficiently fuse data with different fidelity for prediction, and confirmed on the Burgers' equation that it is completely feasible to optimise a high-fidelity model using low-fidelity physical constraints. In order to gain a deeper understanding of MSPCNN's ability to deal with complex phenomena in optimising high-fidelity models using low-fidelity physical constraints, we further conduct shallow water experimental verification. The shallow water equations are a set of hyperbolic partial differential equations that describe the flow below a pressure surface in a fluid, typically water. The governing equations are:

\begin{align}
\frac{\partial h}{\partial t} + \frac{\partial (hu)}{\partial x} + \frac{\partial (hv)}{\partial y} &= 0 \notag \\
\frac{\partial (hu)}{\partial t} + \frac{\partial (hu^2 + \frac{1}{2} g h^2)}{\partial x} + \frac{\partial (huv)}{\partial y} &= 0 \notag \\
\frac{\partial (hv)}{\partial t} + \frac{\partial (huv)}{\partial x} + \frac{\partial (hv^2 + \frac{1}{2} g h^2)}{\partial y} &= 0
\label{eq:Shallow_Water_Equation}
\end{align}

where $h$ is the total water depth (including the undisturbed water depth) with units of meters($m$), $u$ and $v$ are the velocity components in the x (horizontal) and y (vertical) directions with units of meters per second($m/s$), respectively, and $g$ is the gravitational acceleration, typically measured in meters per second squared($m/s^2$). For our simulations, the numerical results are obtained by solving the shallow water equations using a combination of the finite difference method for spatial discretisation and the Euler method for time integration. The high-fidelity data domain is 64×64 grid and the low-fidelity data domain is a 32×32 grid, each containing three channels corresponding to the velocity components $u$, $v$, and the water height $h$. Initial conditions for the simulations involve a cylindrical disturbance in the water height, with the central cylinder's height ranging from $[0.2, 1]$ metres and radius varying between $[4, 16]$ grid units, allowing for a comprehensive study of wave dynamics and fluid behaviour. And the undisturbed water depth is equal to 1 metre.
\subsection{Validation of Multi-Fidelity CAE in Shallow Water Systems}
\begin{figure}
    \begin{center}
        \includegraphics[width=0.8\textwidth]{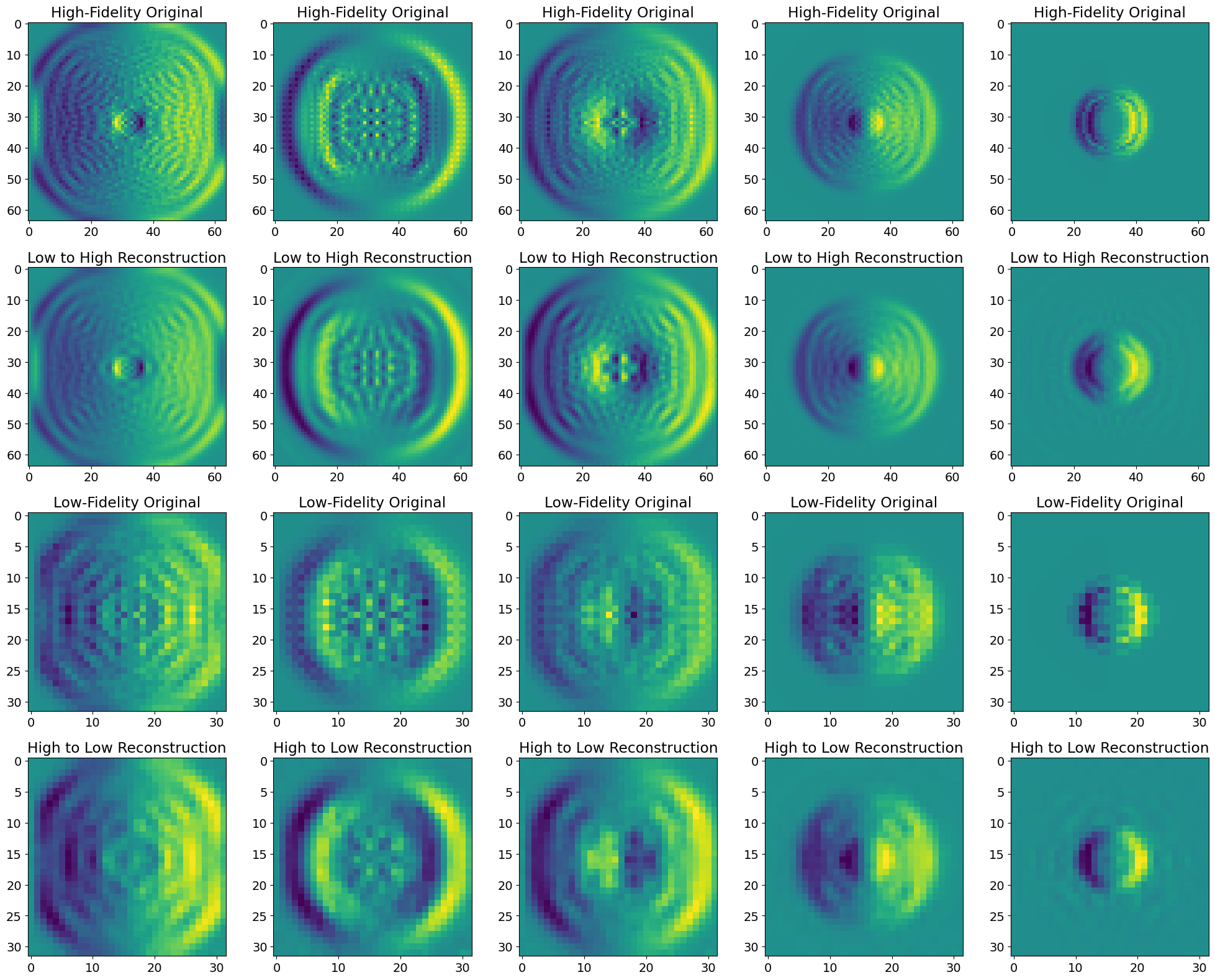}
    \end{center}
    \caption{\label{fig:CAEPrediction} Results from Multi-Fidelity CAE in shallow water systems ($u$ dimension)}
\end{figure}

Similarly, we first showcase the effectiveness of our multi-fidelity CAE in efficiently compressing and then decompressing both high-fidelity and low-fidelity data in Fig.~\ref{fig:CAEPrediction}. We trained the multi-fidelity CAE using 300 corresponding sets of high-fidelity and low-fidelity data. From Fig.~\ref{fig:CAEPrediction}, it's evident that our architecture successfully reconstructs the foundational data for subsequent predictions, demonstrating robust performance across a diverse array of data samples.

\subsection{Effects of Physical Constraints in Shallow Water Systems}

Building upon the validation of our multi-fidelity CAE, we further delve into understanding the role of physical constraints in low-fidelity field in predictions by MSPCNN. For this analysis, we used 300 sets of high-fidelity data as the training set and 30 sets as the test set.

\begin{figure}
    \begin{center}
        \includegraphics[width=1.\textwidth]{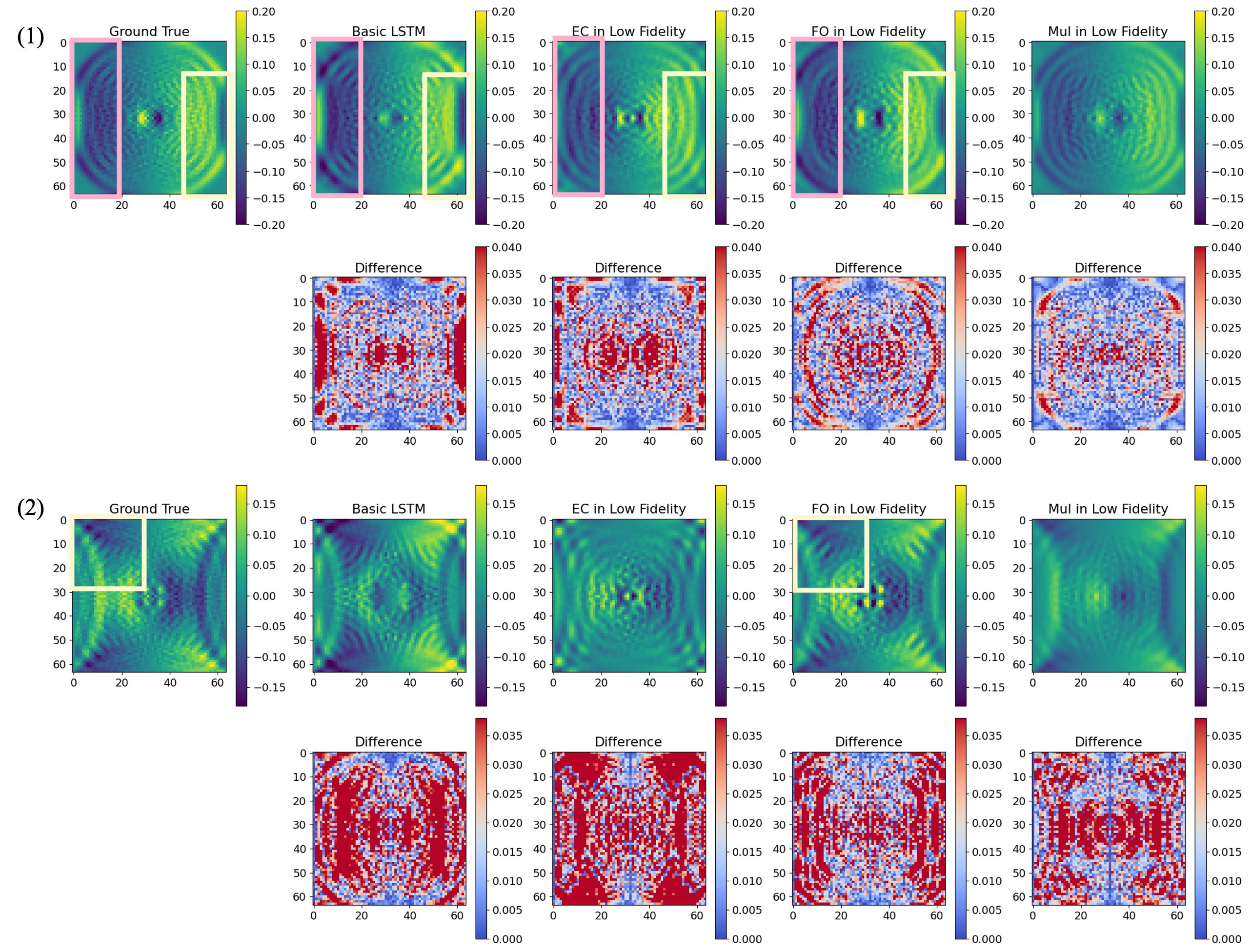}
    \end{center}
    \caption{\label{fig:sw_prediction} Prediction Results ($u$ dimension) and Difference with Groundtrue Comparison of Various LSTM for (1)$t=25$ and (2)$t=120$ in shallow water systems}
\end{figure}

First, we conduct a comparative analysis of various LSTM models based on the shallow water system, as shown in Fig.~\ref{fig:sw_prediction}. In particular, Fig.~\ref{fig:sw_prediction}(1) and (2) are the prediction results and errors comparison of basic LSTM and MSPCNN with various physical constraints in the low-fidelity field at $t=25$ and $t=120$, respectively.
We observe that the basic LSTM model incorrectly captures the evolutionary relationships, resulting in erroneous waveform predictions. Specifically, the model prematurely predicts later-stage waveforms in the early evolution phase ($t=20$), while still incorporating early-stage waveforms during the later evolution phase ($t=120$). This peculiar behavior is highlighted with a pink box in Fig.~\ref{fig:sw_prediction}. This issue persists in MSPCNN that introduces the EC constraint and is alleviated with the embedment of the FO constraint. However, the FO constraint also brings a new issue where the predicted results fail to capture the detailed waveforms as seen in the groundtruth, which is marked with yellow boxes in Fig.~\ref{fig:sw_prediction}. Simultaneously, for the long-time prediction at $t=120$, it is evident that MSPCNN applying the EC constraint causes the prediction results to become slightly smoother, as demonstrated in Fig.~\ref{fig:sw_prediction}(2).

\begin{figure}[]
    \begin{center}
        \includegraphics[width=0.8\textwidth]{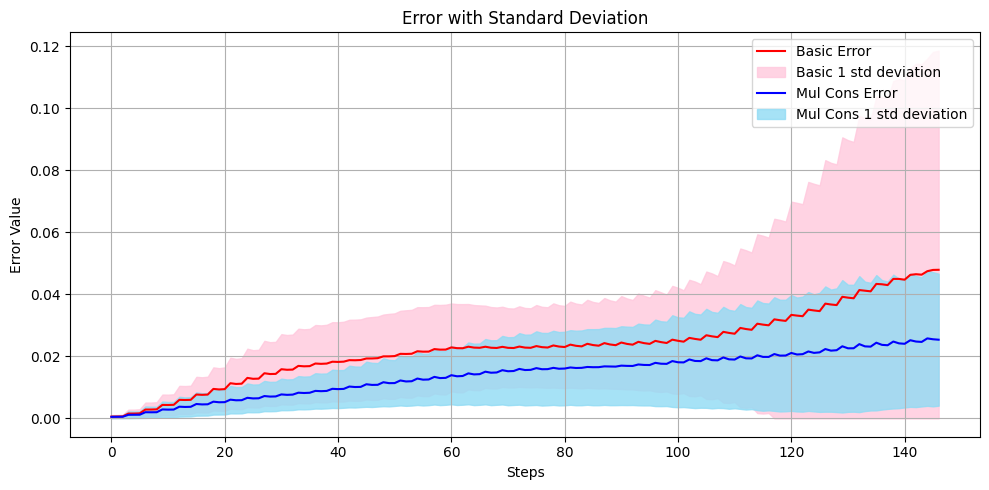}
    \end{center}
    \caption{\label{fig:sw_multi_error} Performance of MSPCNN with Multiple Constraints in Low-Fidelity Field in shallow water systems}
\end{figure}

Furthermore, when we embed both energy conservation and flow operator constraints in the low-fidelity field in MSPCNN, the merits of both constraints are combined to improve the realism of predictions. As shown in Fig.~\ref{fig:sw_prediction}, it improves the clarity and accuracy of early predicted waveforms, making the predicted waveforms less blurred and easier to identify. In addition, multiple constraints also enhance the stability of the model in long-time predictions, alleviating erroneous waveform predictions. However, the employment of the energy constraint still results in smoother predictions, which cannot be completely eliminated. Referring to the metrics detailed in Table~\ref{tab:Performance Comparison between Models SWEs}, the LF-MulCons model achieves an MSE of 53.5\% of the basic model's MSE. This not only marks a significant reduction in prediction error compared to LF-EC and LF-FO, but also underscores the benefits of incorporating various constraints. Relying on various constraints rather than a singular one, proves especially beneficial in intricate systems. Compared with Table~\ref{tab:Performance Comparison between Models Burgers}, it can be easily found that the flow operator has a larger impact on the mse error reduction compared to the application of the energy constraint. We suppose that it might be because flow operators offer more direct influence on fluid behaviour and are effective in capturing complex, nonlinear fluid patterns, leading to precise and nuanced modeling compared to global constraints like energy conservation. In addition, the stability of predictions has also experienced notable enhancements, as indicated by the decreased range of standard deviation depicted in Fig.~\ref{fig:sw_multi_error}.

From the above analysis, when employing MSPCNN to tackle complex physical problems, we conclude that solely relying on a single physical constraint can enhance the authenticity of model predictions to some extent, but it doesn't genuinely improve the prediction accuracy. Combining multiple physical constraints, such as energy conservation and flow operator, can integrate the advantages of different constraints to enhance the realism of model predictions at multiple levels.

\begin{table}[h]
\centering
\caption{Performance Comparison between Models in Shallow Water Systems}
\label{tab:Performance Comparison between Models SWEs}
\begin{tabular}{|c|c|c|c|c|}
\hline
Case & Model & MSE & SSIM & Training Time/Epoch (s) \\
\hline
\multirow{4}{*}{Shallow Water System}     & Basic & 100\% & 0.6497 & 11.38\\
\cline{2-5}
                        & LF-EC & 86.4\% & 0.5166 & 237.61\\
\cline{2-5}
                        & LF-FO & 74.6\% & 0.6277 & 28.30\\
\cline{2-5}
                        & LF-MulCons & 53.5\% & 0.7058 & 256.61\\
\hline
\end{tabular}

\smallskip 
\small{
\begin{tabular}{l}
\textbf{Note:}\\
\textbullet~ Basic: Predictive model trained by solely high-fidelity data.\\
\textbullet~ LF-EC: Model with energy conservation constraint in low-fidelity field.\\
\textbullet~ LF-FO: Model with flow operator constraint in low-fidelity field.\\
\textbullet~ LF-MulCons: Model with multiple constraints in low-fidelity field.\\
\textbullet~ MSE: Mean Squared Error with reference to the basic model set at 100\%.\\
\textbullet~ SSIM: Structural Similarity Index (with data range of 1.0).\\
\textbullet~ Training Time/Epoch (s): Time taken to run one epoch during training, unit: seconds.\\
\textbullet~ \parbox[t]{5in}{The coefficient of the physical constraint $\alpha$ is optimised using the validation set to achieve the best performance for each model. $\alpha_\textrm{EC}$ is the coefficient of energy conservation constraint. $\alpha_\textrm{FO}$ is the coefficient of flow operator constraint. Specifically, $\alpha_\textrm{EC}=4.1e-3$ for LF-EC, $\alpha_\textrm{EC}=1.6e-3$ for LF-MulCons, $\alpha_\textrm{FO}=3.8e-3$ for LF-FO, $\alpha_\textrm{FO}=3.5e-3$ for LF-MulCons.}
\end{tabular}
}

\end{table}

\subsection{Robustness Evaluation with Noisy Data}
In real-world scenarios, particularly when analysing complex systems, models often encounter data that is contaminated with noise. The generation of this noise can result from a multitude of origins, including imprecise measurements, intrinsic uncertainties within the system, or external disturbances. Ensuring the robustness and predictive capabilities of models designed for complex physical systems in the presence of noise is of utmost significance. In order to thoroughly evaluate the stability of our MSPCNN in this particular environment, we conduct a noisy experiment within the shallow water systems. By utilising a model that is trained on data without any noise, we conduct an evaluation of its capacity to make accurate predictions on a dataset that is intentionally contaminated with synthetic noise. This simulation aims to replicate the obstacles encountered in real-world scenarios.

In our experiments, to ensure the representativeness of numerical tests, we utilise spatial correlation patterns that are both homogeneous and isotropic with respect to the spatial Euclidean distance $r=\sqrt{\Delta_x^2+\Delta_y^2}$. This means that they remain unchanged under rotations and translations. We employ these correlation patterns to simulate data errors stemming from various sources. In this context, we consider a Matern type of correlation function~\cite{matern2013spatial}:

\begin{equation}
    \epsilon(r) = (1+\frac{r}{L})\textbf{EXP}(-\frac{r}{L})
\end{equation}
where L is defined as the typical correlation length scale, and we set $L = 4$ for the sake of simplicity.

\begin{figure}
    \begin{center}
        \includegraphics[width=0.8\textwidth]{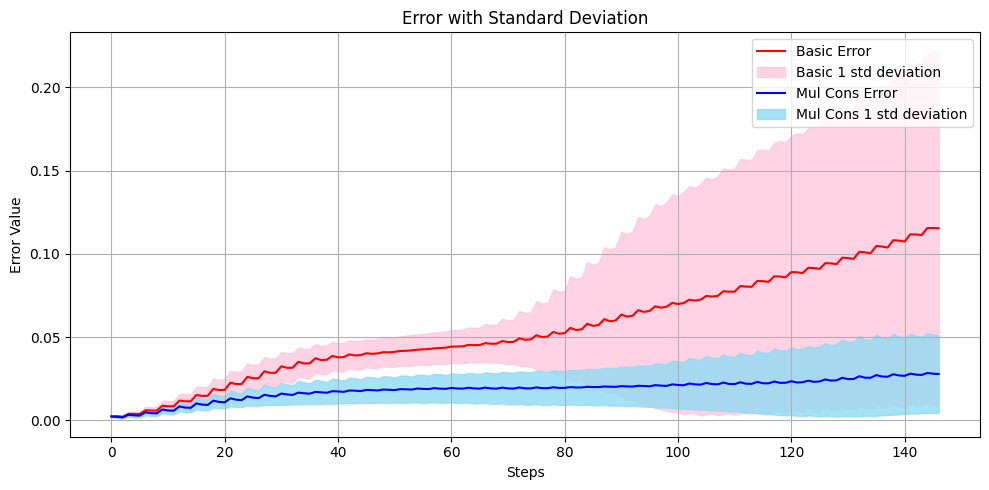}
    \end{center}
    \caption{\label{fig:sw_noise_error} Performance of MSPCNN with Multiple Constraints in Low-Fidelity Field with Noise Initial Condition in shallow water systems}
\end{figure}

In the simulation against noise, we introduce noise into the initial data to obtain the noisy data. This noisy data is then fed into both the basic LSTM and the MSPCNN for recurrent predictions. The outcomes are depicted in Fig.~\ref{fig:sw_noise_error}. When juxtaposed with Fig.~\ref{fig:sw_multi_error}, it's evident that the basic LSTM model struggles with handling noisy data, leading to a remarkably high MSE. Additionally, there's a noticeable expansion in the spread of the standard deviation. In contrast, the MSPCNN fortified with multiple constraints demonstrates resilience against this noise-induced perturbation, registering only a marginal increase in both MSE and the range of the standard deviation. In summary, the MSPCNN demonstrates robust performance when confronted with noisy data.
\section{Conclusion}
\label{sec:conclusion}
Physics-constrained neural networks have emerged as a popular approach for enhancing the reliability of predictions. These networks surpass merely data-driven models by incorporating physical constraint losses into the training process. In this paper, we propose and implement a novel predictive model, MSPCNN. The model is inspired by reducing the cumulative error of long-time prediction while minimising computational cost. Its unique feature is that it can integrate and freely convert data in different fidelities through the multi-fidelity CAE.

We explicitly show that there is significant value in mapping data in various fidelities into a uniform and shared latent space through multi-fidelity CAE. Firstly, it allows low-fidelity data to play a complementary role to high-fidelity data during the training phase as the predictive model accepts latent representations as input. In addition, MSPCNN allows us to enforce physical constraints in the low-fidelity field, instead of applying at a high-fidelity level. As a result, there's a significant reduction in off-line costs, which include expenses related to data acquisition and preprocessing. Meanwhile, this approach guarantees that our model maintains a significant level of accuracy while avoiding the computing challenges commonly encountered by conventional physics-constrained neural networks. While our tests are on a toy model, using this multi-fidelity approach on high-dimensional datasets could offer more significant savings in computation and training costs. Furthermore, the results of shallow water systems emphasise the importance of incorporating multiple constraints while tackling intricate physical problems, since depending exclusively on a solitary constraint may be insufficient. Moreover, the model's adept handling of noisy data highlights its robustness, demonstrating its capacity to provide dependable predictions even in suboptimal circumstances.

The MSPCNN, with its ability to seamlessly encode high- and low-fidelity datasets in a shared latent space and embed physical constraints, offers substantial promise for transforming multiscale simulations in fluid dynamics. Due to its adaptability and computing efficiency, this technology is well-suited for real-time predictive assessments in various areas, including environmental forecasting and industrial fluid operations. Nevertheless, MSPCNN has its limitations. One notable limitation is the error amplification in scenarios with limited spatial correlation, a challenge not unique to MSPCNN but prevalent in traditional models like PCNN. We are addressing this through the development of a custom loss function that better balances simulation fidelity with error reduction. In addition to refining loss functions, another significant avenue for future work is extending our methodology to more complex mesh structures. Currently, both test cases in our study employ squared mesh simulations. However, real-world applications often require modeling on non-structured or even adaptive meshes, where the number and arrangement of meshes can change dynamically to better capture phenomena or optimise computational resources. Furthermore, there's an ongoing exploration to leverage the capabilities of transformer-based models, which can be integrated into the MSPCNN framework as an alternative to traditional CNN and RNN architectures, potentially offering enhanced performance and adaptability.

\section*{Data and code availability}
The code of the burgers equation and the shallow water experiments is available at \url{https://github.com/DL-WG/mspcnn-for-dynamic-system}. Data and the scripts to generate experiments is also provided in the github reporsitory. 

\section*{Acknowledgement}

 This work is supported by the Leverhulme Centre for Wildfires, Environment and Society through the Leverhulme Trust, grant number RC-2018-023 and the EP/T000414/1 PREdictive Modelling with
Quantification of UncERtainty for MultiphasE Systems (PREMIERE).

\section*{Abbreviations}
\noindent
\begin{tabular}{ll}
MSPCNN & Multi-Scale Physics-Constrained Neural Network \\
PDE & Partial Differential Equation \\
CFD & Computational Fluid Dynamics \\
ROM & Reduced Order Modelling \\
ML & Machine Learning \\
DL & Deep Learning \\
AE & Autoencoder \\
RNN &  Recurrent Neural Network \\
LSTM &  Long Short-Term Memory \\
CAE &  Convolutional Autoencoder \\
CNN & Convolutional Neural Network \\
MSE &  Mean Square Error \\
RMSE &  Root Mean Square Error \\
PCNN &   Physics-Constrained Neural Network \\
MAE &  Mean Absolute Error \\
EC &  Energy Conservation \\
FO &  Flow Operator \\
SSIM &  Structural Similarity Index \\
HF-EC & Model with energy conservation constraint in high-fidelity field \\
LF-EC & Model with energy conservation constraint in low-fidelity field \\
HF-FO & Model with flow operator constraint in high-fidelity field \\
LF-FO & Model with flow operator constraint in low-fidelity field \\
HF-MulCons & Model with multiple constraints in high-fidelity field \\
LF-MulCons & Model with multiple constraints in low-fidelity field \\
\end{tabular}

\newpage
\footnotesize
\bibliographystyle{elsarticle-num-names}
\bibliography{references}  

\end{document}